\PassOptionsToPackage{table}{xcolor}
\documentclass[onecolumn]{dataflow}
\usepackage{adjustbox}
\newif\ifPoViTableFloat
\AtBeginEnvironment{table}{\PoViTableFloattrue}
\AtBeginEnvironment{table*}{\PoViTableFloattrue}
\let\PoViNativeTabular\tabular
\let\PoViNativeEndTabular\endtabular
\newsavebox{\PoViZeroTableBox}
\newsavebox{\PoViUnitTableBox}
\newsavebox{\PoViFullWidthTableBox}

\ExplSyntaxOn
\tl_new:N \l__povi_table_columns_tl
\dim_new:N \l__povi_table_zero_dim
\dim_new:N \l__povi_table_step_dim

\cs_new_protected:Npn \__povi_native_table:nnn #1#2#3
  {
    \PoViNativeTabular[#1]{#2}
    #3
    \PoViNativeEndTabular
  }

\cs_new_protected:Npn \__povi_fixed_table:nnn #1#2#3
  {
    \group_begin:
      \cs_set_eq:NN \PoViSavedMulticolumn \multicolumn
      \cs_set:Npn \multicolumn ##1##2##3
        { \PoViSavedMulticolumn{##1}{##2}{} }
      \dim_zero:N \tabcolsep
      \sbox{\PoViZeroTableBox}{\__povi_native_table:nnn {#1}{#2}{#3}}
      \dim_set:Nn \tabcolsep {1pt}
      \sbox{\PoViUnitTableBox}{\__povi_native_table:nnn {#1}{#2}{#3}}
      \cs_set_eq:NN \multicolumn \PoViSavedMulticolumn

      \dim_set:Nn \l__povi_table_zero_dim {\box_wd:N \PoViZeroTableBox}
      \dim_set:Nn \l__povi_table_step_dim
        {\box_wd:N \PoViUnitTableBox - \l__povi_table_zero_dim}
      \dim_set:Nn \tabcolsep
        {
          \fp_to_dim:n
            {
              max(0,
                (\dim_to_fp:n{\textwidth}
                  - \dim_to_fp:n{\l__povi_table_zero_dim})
                / \dim_to_fp:n{\l__povi_table_step_dim}
              )
            }
        }
      \sbox{\PoViFullWidthTableBox}{\__povi_native_table:nnn {#1}{#2}{#3}}
      \typeout{PoVi~table~\thetable\c_colon_str~width=\the\wd\PoViFullWidthTableBox;~textwidth=\the\textwidth;~padding=\the\tabcolsep}
      \clipbox{0pt~0pt~0pt~0pt}{\usebox{\PoViFullWidthTableBox}}
    \group_end:
  }

\NewDocumentCommand{\PoViFullWidthTable}{m m +m}
  {
    \tl_set:Nn \l__povi_table_columns_tl {#2}
    \tl_if_eq:nnT {#2} {@{}llcp{6.6cm}l@{}}
      {\tl_set:Nn \l__povi_table_columns_tl {llcp{6.6cm}l}}
    \tl_if_eq:nnT {#2} {@{}lcccc@{}}
      {\tl_set:Nn \l__povi_table_columns_tl {lcccc}}
    \tl_if_eq:nnT {#2} {@{}lcc@{}}
      {\tl_set:Nn \l__povi_table_columns_tl {lcc}}
    \tl_if_eq:nnT {#2} {lcccccccc@{\hskip8pt}cccc}
      {\tl_set:Nn \l__povi_table_columns_tl {lcccccccccccc}}
    \exp_args:NnV \__povi_fixed_table:nnn {#1} \l__povi_table_columns_tl {#3}
  }
\ExplSyntaxOff

\RenewDocumentEnvironment{tabular}{O{c} m +b}{%
  \ifPoViTableFloat
    \PoViFullWidthTable{#1}{#2}{#3}%
  \else
    \begin{adjustbox}{max width=\linewidth}%
      \PoViNativeTabular[#1]{#2}%
      #3%
      \PoViNativeEndTabular
    \end{adjustbox}%
  \fi
}{}
\BeforeBeginEnvironment{tabular*}{\begin{adjustbox}{max width=\linewidth}}
\AfterEndEnvironment{tabular*}{\end{adjustbox}}

\newif\ifPoViSingleColumnFigure
\AtBeginEnvironment{figure}{%
  \PoViSingleColumnFiguretrue
  \captionsetup{width=\textwidth}%
}
\AtBeginEnvironment{figure*}{%
  \PoViSingleColumnFigurefalse
  \captionsetup{width=\textwidth}%
}
\let\PoViUnscaledIncludeGraphics\includegraphics

\NewDocumentCommand{\PoViDrawOriginalGraphics}{m m m}{%
  \IfBooleanTF{#1}{%
    \PoViUnscaledIncludeGraphics*[#2]{#3}%
  }{%
    \PoViUnscaledIncludeGraphics[#2]{#3}%
  }%
}

\RenewDocumentCommand{\includegraphics}{s O{} m}{%
  \ifPoViSingleColumnFigure
    \ifdim\linewidth=\textwidth
      \typeout{PoVi single-column graphic: #3; width=\the\textwidth}%
      \ifstrempty{#2}{%
        \PoViDrawOriginalGraphics{#1}{width=\textwidth}{#3}%
      }{%
        \PoViDrawOriginalGraphics{#1}{#2,width=\textwidth}{#3}%
      }%
    \else
      \PoViDrawOriginalGraphics{#1}{#2}{#3}%
    \fi
  \else
    \PoViDrawOriginalGraphics{#1}{#2}{#3}%
  \fi
}

\renewcommand{\titlefont}{%
  \sffamily\fontsize{16}{20}\selectfont\raggedright
  \hyphenpenalty=10000\relax\exhyphenpenalty=10000\relax
}

\DeclareFontShape{T1}{bytesans}{m}{it}{<->ssub*lmss/m/sl}{}
\DeclareFontShape{T1}{bytesans}{b}{it}{<->ssub*lmss/bx/sl}{}
\DeclareFontShape{T1}{bytesans}{bx}{it}{<->ssub*lmss/bx/sl}{}

\renewcommand\cite{\citep}

\renewcommand{\templatelogo}{%
  \includegraphics[height=9mm]{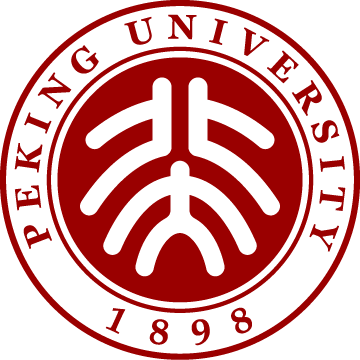}%
  \hspace{3mm}%
  \includegraphics[height=9mm]{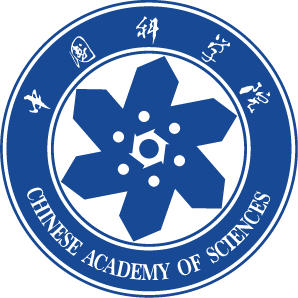}%
  \hspace{4mm}%
  \includegraphics[height=9mm]{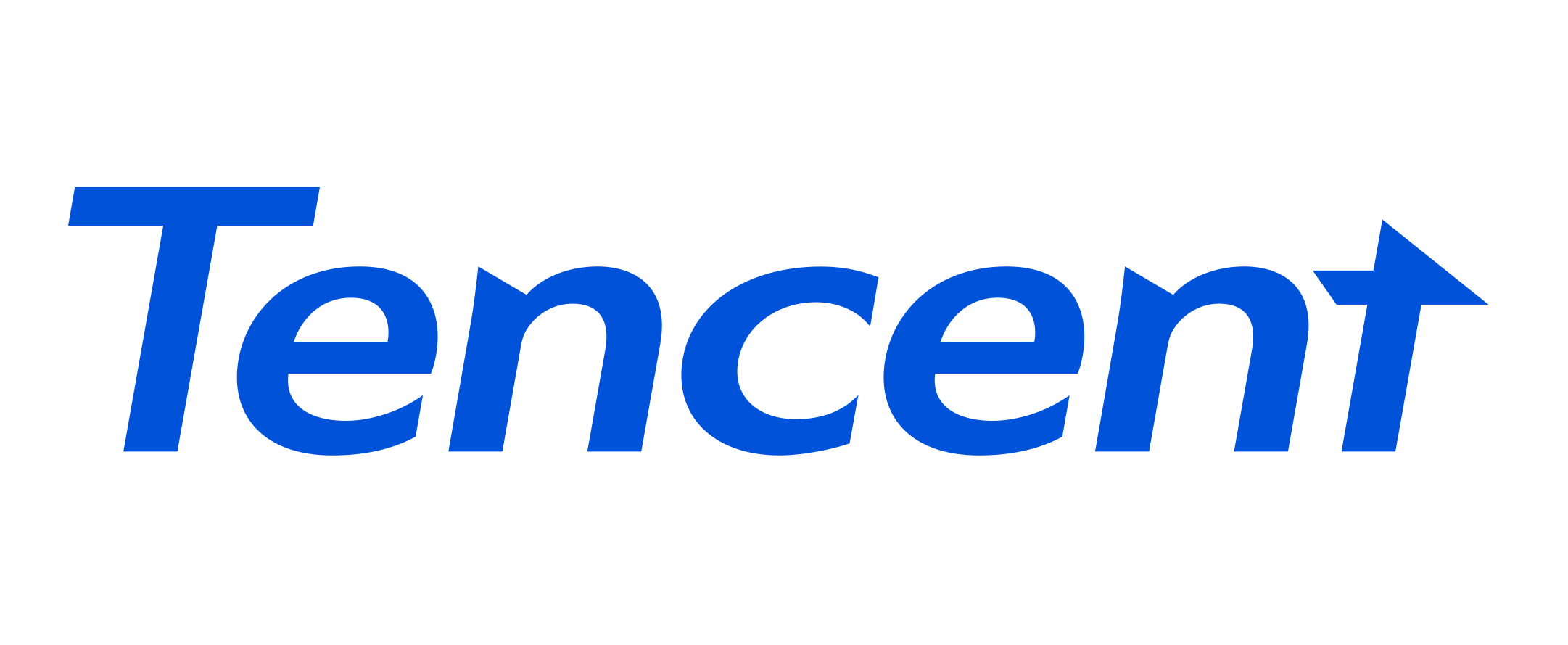}%
}
\fancypagestyle{firststyle}{%
  \fancyhf{}
  \renewcommand{\headrulewidth}{0pt}
  \fancyhead[L]{\templatelogo}
  \fancyhead[R]{%
    \raisebox{2.5mm}{\normalfont\small\textcolor[HTML]{4D4D4D}{\today}}%
  }
}
\let\PoViTemplateAuthor\author
\renewcommand{\author}[1]{%
  \gdef\PoViOriginalAuthorMetadata{#1}%
  \begingroup
    \let\author\PoViTemplateAuthor
\author[1,*]{Yuxia Zhu}
\author[3,*]{Shiying Zhang}
\author[2,*]{Lingzhuang Sun}
\author[3]{Song Yang}
\author[1]{Siheng Wu}
\author[2]{Xinglong Xu}
\author[2]{Sizhe Qiu}
\author[1]{Hao Liang}
\author[4,\dagger]{Cheng Tan}
\author[4,\dagger]{Asher Qin}
\author[1,\dagger]{Wentao Zhang}

\affiliation[]{%
  $^{1}$Peking University\quad
  $^{2}$University of Chinese Academy of Sciences\\
  $^{3}$Tongji University\quad
  $^{4}$Tencent
}
\contribution[*]{Equal contribution.}
\contribution[\dagger]{Corresponding author.}%
  \endgroup
}

\let\PoViTemplateAbstract\abstract
\let\PoViTemplateMakeTitle\maketitle
\renewcommand{\maketitle}{}
\RenewDocumentEnvironment{abstract}{+b}{%
  \PoViTemplateAbstract{#1}%
  \PoViTemplateMakeTitle
}{}

\RenewDocumentCommand{\addtolist}{O{}+m+m+m+m}{%
  \ifdefempty{#3}{}{\gappto#3{#5}}%
  \gappto#3{#4[#1]{#2}}%
}

\renewcommand{\email}[1]{%
  \href{mailto:#1}{\textcolor{black}{\normalfont\small\detokenize{#1}}}%
}

\renewcommand{\checkdataformat}[2][]{%
  \par\noindent
  \parbox[t]{55pt}{%
    \checkdatafont\sffamily\bfseries\strut\mbox{#1:}}%
  \parbox[t]{\dimexpr\linewidth-55pt\relax}{%
    \normalfont\small\raggedright\strut #2}\par
}

\makeatletter
\renewcommand{\mymaketitle}{%
  \vspace*{-4mm}%
  \@toptitlebar
  \begin{tcolorbox}[
    enhanced,colback=boxbg,frame hidden,arc=3mm,
    left=8mm,right=8mm,top=5mm,bottom=5mm,
    before skip=0pt,after skip=0pt
  ]
    \setlength{\parskip}{0pt}
    \setlength{\parindent}{0pt}
    {\titlelist\par}
    \vskip 3mm
    {\centering\authorlist\par}
    \vskip 2mm
    {\centering\affiliationlist\par}
    \vskip 3mm
    \abstractlist\par
    \ifdefempty{\contributionlist}{}{%
      \vskip 2mm
      \contributionlist\par
    }
    \ifdefempty{\checkdatalist}{}{%
      \vskip 2mm
      \checkdatalist\par
    }
  \end{tcolorbox}
  \tcbset{reset}
  \FloatBarrier
  \thispagestyle{firststyle}
}
\makeatother
\renewcommand{\PoViTemplateMakeTitle}{\mymaketitle\vskip 4mm}

\newcommand{\emailicon}{\raisebox{-1.5pt}{\includegraphics[height=1.05em]{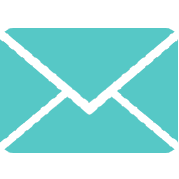}}}

\newcommand{\sourcelink}
{\raisebox{-1.5pt}{\includegraphics[height=1.05em]{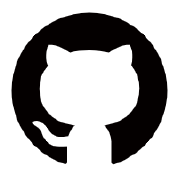}}}

\newcommand{\websitelink}
{\raisebox{-1.5pt}{\includegraphics[height=1.05em]{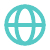}}}

\newcommand{\contactlabel}[2]{%
  \makebox[11pt][c]{#1}\hspace{3pt}#2}

\checkdata[\contactlabel{\emailicon}{Email}]{%
  \email{Zzzyxiia@outlook.com}}

\checkdata[\contactlabel{\sourcelink}{Code}]{%
  \href{https://github.com/OpenRaiser/PoVi}
       {https://github.com/OpenRaiser/PoVi}}

\checkdata[\contactlabel{\websitelink}{Website}]{%
  \href{https://openraiser.github.io/PoVi}
       {https://openraiser.github.io/PoVi}}
\usepackage{enumitem}
\usepackage{latexsym}

\usepackage[T1]{fontenc}
\usepackage[utf8]{inputenc}

\usepackage{microtype}

\usepackage{graphicx}

\usepackage{booktabs}
\usepackage{multirow}
\usepackage{amsmath}
\usepackage{tabularx}
\usepackage{placeins}

\usepackage[table]{xcolor}

\definecolor{bestblue}{RGB}{226,238,249}
\newcommand{\best}[1]{%
  \cellcolor{bestblue}\textbf{#1}%
}

\usepackage[most]{tcolorbox}
\usepackage{enumitem}

\newcommand{\promptfield}[1]{\texttt{\detokenize{#1}}}
\newcommand{\promptlabel}[1]{%
  {\color{orange!70!black}\bfseries\textsc{#1}}\enspace}

\newtcolorbox{promptcard}[2][]{
  enhanced,
  breakable,
  colback=white,
  colframe=black!22,
  colbacktitle=orange!7,
  coltitle=black,
  title={#2},
  fonttitle=\bfseries\small,
  fontupper=\footnotesize,
  fontlower=\footnotesize,
  boxrule=0.45pt,
  arc=0.8mm,
  left=1.8mm,
  right=1.8mm,
  top=1.4mm,
  bottom=1.4mm,
  titlerule=0.35pt,
  borderline west={1.8pt}{0pt}{orange!65!black},
  before skip=7pt,
  after skip=9pt,
  #1
}

\usepackage{tikz}

\title{PoVi: Agentic Poetry-to-Video Generation with Knowledge Grounding and Figurative Understanding}

\author{First Author \\
  Affiliation / Address line 1 \\
  Affiliation / Address line 2 \\
  Affiliation / Address line 3 \\
  \texttt{email@domain} \\\And
  Second Author \\
  Affiliation / Address line 1 \\
  Affiliation / Address line 2 \\
  Affiliation / Address line 3 \\
  \texttt{email@domain} \\}

\begin{document}
\maketitle
\begin{abstract}
Poetry-to-video generation requires more than translating text into visual scenes. Poetic language is highly compressed, often depends on historical and cultural knowledge, and frequently conveys meaning through metaphor, imagery, and other non-literal expressions. We introduce PoVi, a benchmark and agentic framework for grounded poetry-to-video generation, evaluating semantic grounding, discourse coherence, and audiovisual realization. PoVi reconstructs poem semantics into an evidence-bounded representation and progressively transforms them into narration, interpretation, character assets, and shot-level visual plans, followed by narration-aligned video generation. Across three planning backbones and four video generators, PoVi consistently improves over single-call generation, with the largest gains on implicit background knowledge and metaphorical relations. PoVi provides a testbed for studying knowledge-grounded, long-horizon multimodal generation beyond direct text-to-video synthesis.
\end{abstract}

\section{Introduction}

Recent video generation models have made remarkable progress in visual fidelity, temporal consistency, and controllability~\cite{blattmann2023stable,chen2024videocrafter2,yang2025cogvideox,bar2024lumiere}. Modern systems can synthesize complex motion~\cite{kondratyuk2023videopoet,zhang2025tora,polyak2024movie}, maintain subjects across shots~\cite{gao2025seedance,seedance2026seedance,seedance25_2026,zhou2024storydiffusion}, and increasingly support multimodal references~\cite{wang2023videocomposer,jiang2025vace} and filmmaking-oriented controls. Yet these advances largely assume that the input already specifies, at least approximately, what should appear on screen~\cite{tian2024videotetris}, leaving the main challenge as faithfully generating the specified content.

Classical poetry poses a fundamentally different problem. A poem is not a visual specification, but a highly compressed literary representation in which characters, settings, causal relations, historical background, and even the referents of salient imagery may remain unstated~\cite{chen2025benchmarking,chiu2025culturalbench}. A few lines may presuppose an episode from a poet's life, compress several events into a single poetic image, or use a concrete object as the source of a metaphor whose target is never explicitly named~\cite{tong2024metaphor,sanchez2025metaphor}. Faithful interpretation therefore requires both \textit{knowledge grounding}, to recover historical and cultural context beyond the poem itself, and \textit{figurative understanding}, to resolve metaphor, imagery, and other non-literal expressions within it. Directly prompting a video generator with the poem may therefore produce visually plausible but interpretively shallow results: surface imagery is rendered, while the background, narrative progression, or figurative meaning is lost~\cite{mao2024metapro,jiang2024poetry2image}. Poetry-to-video generation thus requires first determining \emph{what should be rendered}, before deciding \emph{how} to render it.

\begin{figure*}[ht]
  % % \vspace{-2mm}
  \includegraphics[width=\textwidth]{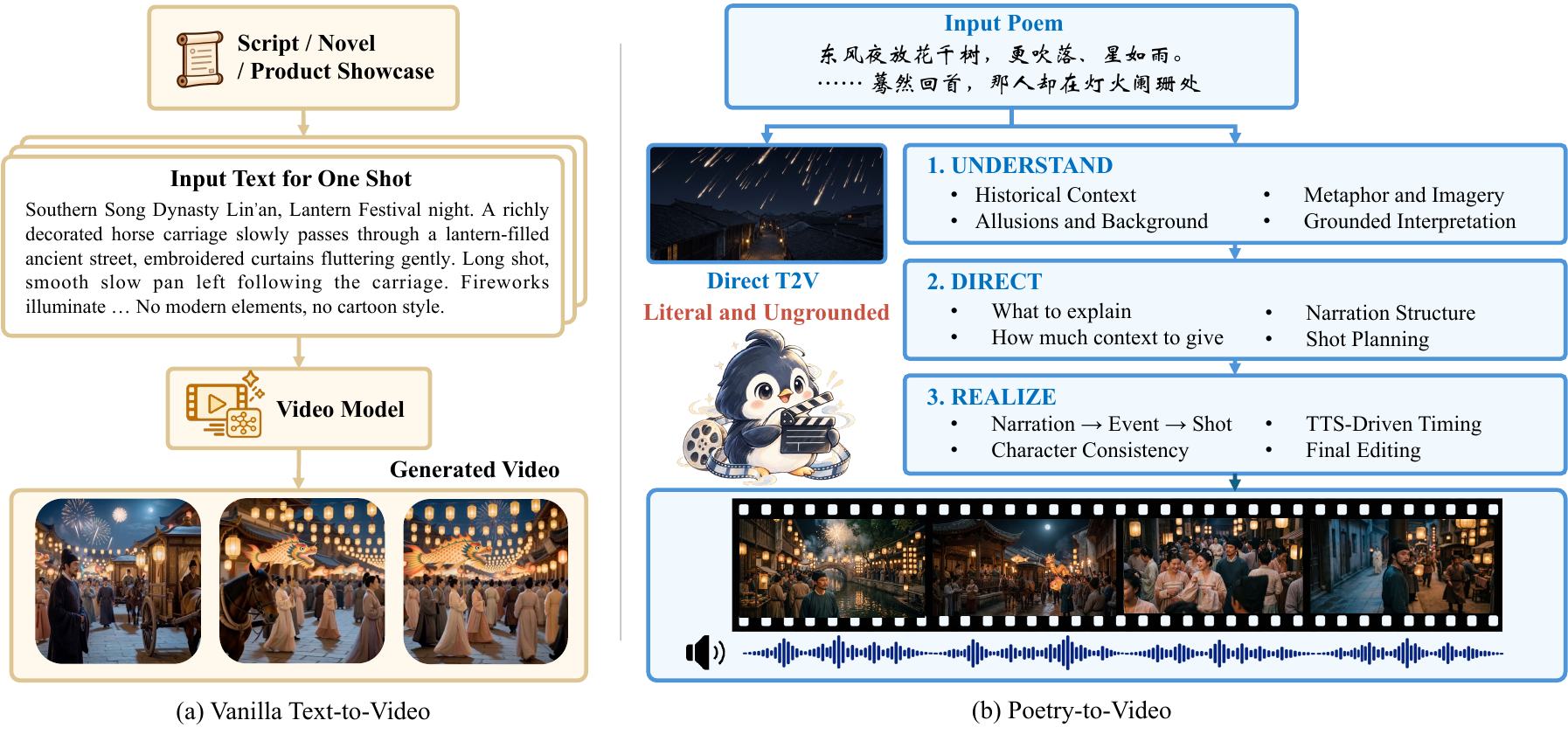}
  % \vspace{-2mm}
  \caption{Comparison between vanilla text-to-video generation and poetry-to-video generation. Unlike ordinary text prompts, poetry requires understanding implicit meaning, directing how that meaning should be communicated, and realizing the resulting plan as aligned audiovisual content.}
  % \vspace{-2mm}
  \label{fig:introduction}
\end{figure*}

Recent agentic video generation systems organize video creation through stories, scripts, storyboards, and shot plans, but typically assume that the intended content is already sufficiently specified. Poetry introduces an additional challenge: before directing can begin, the system must first recover implicit characters, events, historical context, and figurative relations, and determine what information should be conveyed. It must then organize this understanding into a coherent narration and visual plan before realizing it as synchronized audiovisual content. Errors at the understanding stage cannot be corrected by a stronger video generator if the intended meaning has never been made explicit. As illustrated in Figure~\ref{fig:introduction}, poetry-to-video therefore requires a progression from \emph{understanding}, through \emph{directing}, to \emph{audiovisual realization}.

In this work, we introduce \textbf{PoVi}, a benchmark and agentic framework
for grounded poetry-to-video generation. PoVi-Bench comprises 120 classical
Chinese works, evenly divided between 60 Tang poems and 60 Song \emph{ci}.
Within each literary form, the benchmark contains 20 easy, 20 medium, and
20 challenging works. Each work is accompanied by structured annotations
covering required information, metaphor mappings, evidential support,
ordered background events, and visual requirements. Building on these
annotations, we develop a comprehensive suite of metrics that evaluates
semantic grounding, discourse coherence, and audiovisual realization.

We further propose an \textit{evidence bounded multi agent framework} in which a knowledge planning agent constructs a semantic director graph that organizes historical context, causal relations, metaphorical mappings, and visual constraints with supporting evidence. Role specialized agents then convert this representation into narration, interpretation, character assets, and shot-level visual plans. PoVi further uses measured narration duration to align clips, speech, subtitles, and transitions. Across three planning backbones and four video generators, PoVi consistently outperforms single call generation. The largest gains occur on implicit background knowledge and metaphorical relations, showing that poetry to video generation depends on recovering and organizing the meaning that the video should convey.

Our main contributions are as follow:
\begin{itemize}
\vspace{-2mm}
\setlength{\itemsep}{1pt}
\setlength{\parskip}{0pt}
\setlength{\parsep}{0pt}
\item We formulate \textbf{poetry-to-video} as grounded multimodal generation and introduce PoVi-Bench for comprehensive evaluation.
\item We propose an evidence-bounded multi-agent framework for semantic reconstruction, directing, and narration-aligned video generation.
\item PoVi consistently improves over single-call generation, especially on background knowledge and metaphorical relations.
\end{itemize}

\section{Related Works}

\subsection{Text-to-Video Generation}

Recent advances in text-to-video generation have greatly improved visual fidelity, temporal coherence, and overall controllability. Models such as HunyuanVideo~\cite{kong2024hunyuanvideo,wu2025hunyuanvideo}, Wan~\cite{wan2025wan}, CogVideoX~\cite{yang2025cogvideox}, Veo~\cite{wiedemer2025video}, Seedance~\cite{seedance2026seedance,seedance25_2026}, and MiniMax H3~\cite{minimax-website} support increasingly complex, coherent, and controllable video generation. However, conventional T2V largely assumes that the input already specifies what should appear on screen. Classical poetry violates this assumption, as key characters, events, contexts, and figurative meanings may remain implicit. Poetry-to-video therefore requires first recovering \emph{what should be rendered} before determining \emph{how to render it}.

\subsection{Poetry Understanding and Generation}
Classical Chinese poetry has been extensively studied in language generation and understanding due to its compressed semantics, formal constraints, cultural references, and historical allusions. Recent LLM-based systems such as CharPoet~\cite{yu2024charpoet} and PoeTone~\cite{qu2026poetone} explore controlled poetry generation, while Jiuge-Tuiqiao~\cite{han2026jiuge} incorporates classical encyclopedic knowledge and historical poetic evidence for knowledge-grounded and interpretable refinement. Nevertheless, most existing work remains within the text-to-text paradigm, focusing on generation, revision, appreciation, or explanation.

Recent efforts such as Poetry2Image~\cite{jiang2024poetry2image} use iterative LLM-based correction to improve the completeness and semantic consistency of poetry-conditioned images, while SE2Image~\cite{wang2025se2image} separates scene and emotion descriptions to guide image generation. These works highlight the need for semantic grounding before visual generation, but remain limited to static outputs. Poetry-to-video additionally requires temporal and narrative planning, including event ordering, scene transitions, and coordination between narration and visual content.

\subsection{Agentic Long-form Video Generation}
Recent work has shifted long-form video generation toward structured storytelling and agentic filmmaking pipelines~\cite{hu2024storyagent}. StoryDiffusion~\cite{zhou2024storydiffusion}, VideoDirectorGPT~\cite{lin2023videodirectorgpt}, VideoGen-of-Thought~\cite{zheng2024videogen}, ShotAdapter~\cite{kara2025shotadapter}, and STAGE~\cite{zhang2026stage} explore story-, script-, storyboard-, and shot-level representations for coherent multi-shot generation. VideoStudio~\cite{long2024videostudio}, ViMax~\cite{huang2026vimax}, and OpenMontage~\cite{openmontage2026} further introduce agentic planning and production, while FilmWorld~\cite{zuo2026filmworld} incorporates persistent cinematic world modeling.

Poetry introduces an additional stage before directing: the system must first recover implicit characters, events, emotions, and cultural context, then decide how this meaning should be distributed across narration and visuals. PoVi therefore extends agentic filmmaking with an explicit semantic reconstruction stage that bridges literary interpretation and downstream video generation.

\section{Preliminaries}
\label{sec:preliminaries}

Given a poem record \(x=(\mathcal{L},m,B)\), where
\(\mathcal{L}=\langle\ell_1,\ldots,\ell_n\rangle\) is the ordered sequence
of poem lines, \(m\) contains metadata such as title, author, dynasty, and
literary form, and \(B\) denotes optional external reference material available
at inference time, the task is to produce a narrated explanatory video
\(y=(V,A,T)\). Here, \(V\) is the rendered visual track and \(A\) is the aligned narration track. We allow \(B=\emptyset\), corresponding to the setting in which no external reference is provided.

Unlike conventional text-to-video generation, \(x\) is not a visual
specification: characters, settings, event relations, historical context,
and figurative referents may remain implicit and must be recovered before
visual decisions can be made. We evaluate systems purely through their
observable outputs, including the narration text, rendered video, and audio
with a time-aligned transcript; no access to internal representations is
required. This interface allows single-pass and agentic systems to be
evaluated under the same output format.

\section{Benchmark}
\label{sec:benchmark}
% We instantiate the task defined in Section~\ref{sec:preliminaries} through
% PoVi-Bench, a balanced and fully annotated benchmark of 120 classical Chinese
% works.

\subsection{Benchmark Composition}
\label{ssec:composition}

Table~\ref{tab:eval-composition} summarizes the composition of PoVi-Bench, which contains 120 works, evenly divided between 60 Tang poems and 60 Song \emph{ci}. Within each literary form, 20 works are assigned to each difficulty level. Median length increases from 35 characters for easy works to 56 for medium and 136 for challenging ones. Difficulty also reflects linguistic accessibility, allusion density, structural complexity, and interpretive ambiguity.

\begin{table}[ht]
\centering
\small
\setlength{\tabcolsep}{6pt}
\caption{Composition of PoVi-Bench across literary forms and difficulty levels.}
\label{tab:eval-composition}
\begin{tabular}{lrrrr}
\toprule
Literary form & Easy & Medium & Challenging & Total \\
\midrule
Tang poetry & 20 & 20 & 20 & 60 \\
Song ci     & 20 & 20 & 20 & 60 \\
\midrule
\textbf{Total} & \textbf{40} & \textbf{40} & \textbf{40} & \textbf{120} \\
\bottomrule
\end{tabular}
\end{table}

% Difficulty also corresponds to a clear increase in textual complexity:
% the median poem length rises from 35 characters for easy examples to 56
% for medium and 136 for challenging ones. Difficulty, however, is not
% determined by length alone, but also considers linguistic accessibility,
% allusion density, structural complexity, and interpretive ambiguity.

% \begin{figure*}[ht]
% \centering
% \includegraphics[width=\textwidth]{figures/povi_framework.pdf}
% \caption{The PoVi framework. A knowledge planner constructs and validates a
% semantic director graph; five role-specialized stages realize the background
% story, narration, close reading, character assets, and visual direction;
% measured speech durations then control media generation, edit groups,
% subtitles, and final assembly.}
% \label{fig:pipeline}
% \end{figure*}

\subsection{Gold Annotations}
\label{ssec:gold}

All 120 works are exhaustively annotated with frozen gold records
containing five field groups:

\begin{itemize}[leftmargin=1.2em,itemsep=1pt,topsep=3pt,parsep=0pt]
\item \texttt{required\_information}: facts and explanations a faithful account must convey, each with an importance weight.
\item \texttt{metaphor\_maps}: source/target domain, and the bridges licensing each figurative expression.
\item \texttt{evidence\_claims}: the textual, historical, or scholarly support admissible for each reading.
\item \texttt{key\_events}: ordered background events, used for both narrative and temporal scoring.
\item \texttt{visual\_requirements}: visual content the video must realize, together with its \emph{forbidden realizations}.
\end{itemize}
Every metaphor map also carries an interpretive status: \textsc{consensus},
\textsc{reasonable}, \textsc{disputed}, or \textsc{unknown}, which lets us
check whether a system presents a disputed reading as fact.

\subsection{Metric Suite}
\label{ssec:metrics}

\begin{figure}[t]
    \centering
    \includegraphics[width=\columnwidth]{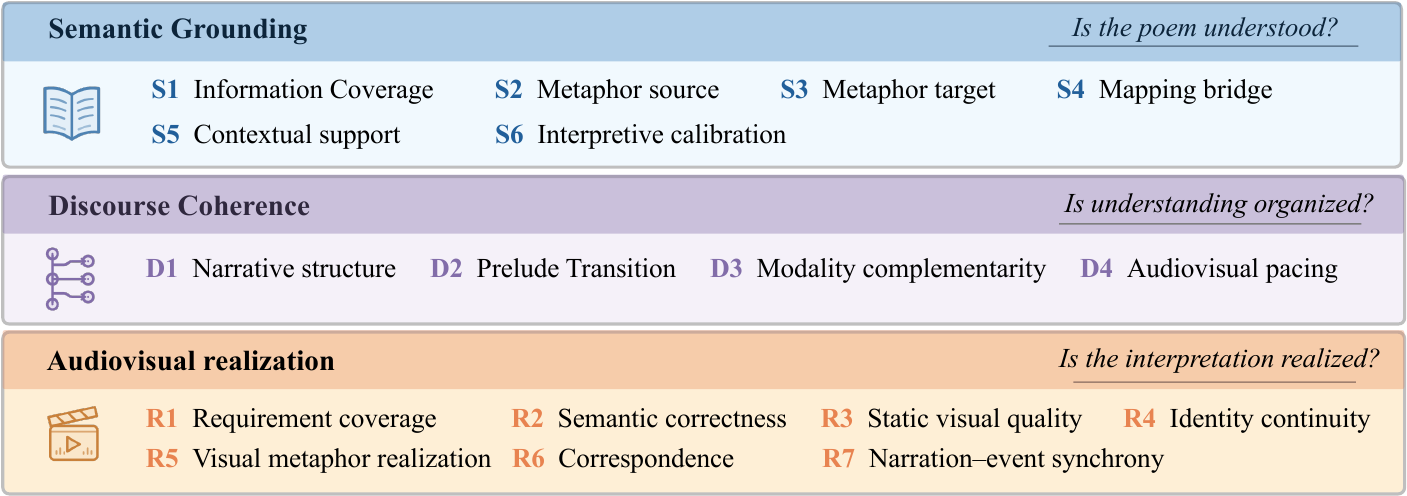}
    \caption{Overview of the 17 PoVi evaluation metrics, grouped into semantic grounding, discourse coherence, and audiovisual realization.}
    \vspace{-4mm}
    \label{fig:metrics}
\end{figure}

We define 17 primary metrics across three dimensions: semantic grounding, discourse coherence, and audiovisual realization (Figure~\ref{fig:metrics}). Coverage and correctness metrics are scored in $[0,1]$, while quality metrics use a 1--5 rubric and are normalized.
% when reported together.

Coverage-based metrics aggregate scores over frozen gold items, while quality metrics are evaluated over the relevant narration or video span. R1 and R2 separate coverage from correctness, and metaphor is evaluated in both narration (S2--S4) and video (R5). Full definitions and aggregation rules are given in Appendix~\ref{app:metrics} (Table~\ref{tab:metrics}).

\paragraph{Scoring Protocol}
\label{ssec:protocol}

Gold annotations are frozen before evaluation, and each judge request scores one atomic metric. For video-based metrics, evidence is collected over sliding temporal windows and each gold item is adjudicated once using all relevant evidence.
% , avoiding repeated penalties for content that appears later in the video.

\paragraph{Judge Meta-Evaluation}
\label{ssec:meta}

Primary results use human ratings. We also evaluate all outputs with
Gemini-3.5-Flash under the same metrics and frozen annotations, and measure
its agreement with human ratings over all 120 works. Per-condition automatic scores and human--Gemini
agreement are reported in Appendix~\ref{app:full}.

\section{The PoVi Framework}
\label{sec:method}

\subsection{Overview}

PoVi factorizes poetry-to-video generation into semantic planning, structured
realization, and audiovisual assembly:
% \[
\begin{equation}    
x \;\xrightarrow{\;\Pi\;}\; G
  \;\xrightarrow{\;\Phi\;}\; \pi
  \;\xrightarrow{\;\Psi\;}\; y .
\end{equation}
% \]
The planning stage \(\Pi\) constructs a semantic director graph \(G\);
the realization stage \(\Phi\) converts \(G\) into a structured text and
shot plan \(\pi\); and the assembly stage \(\Psi\) generates media assets
and assembles the final video \(y\) under deterministic timing constraints.

\subsection{Evidence-Bounded Planning}

\paragraph{Evidence.}
We construct an evidence set $E$: the poem and metadata provide
medium-authority evidence, scholarly references in $B$ high-authority evidence,
and model-recalled background knowledge low-authority evidence. When $B$ is
absent, recalled biographical, historical, and interpretive claims are recorded by the model and frozen for downstream citation.
% We first build an evidence set \(E\). The poem and metadata are treated as medium-authority evidence, provided scholarly references in \(B\) as high-authority evidence, and model-recalled background knowledge as low-authority evidence. When external background material is absent, the model writes an inventory of what it recalls about the poet, including biographical events, historical conditions, the writing occasion, and interpretive claims, which are then frozen so that later stages may cite but not extend them.

\begin{figure*}[ht]
\centering
\includegraphics[width=\textwidth]{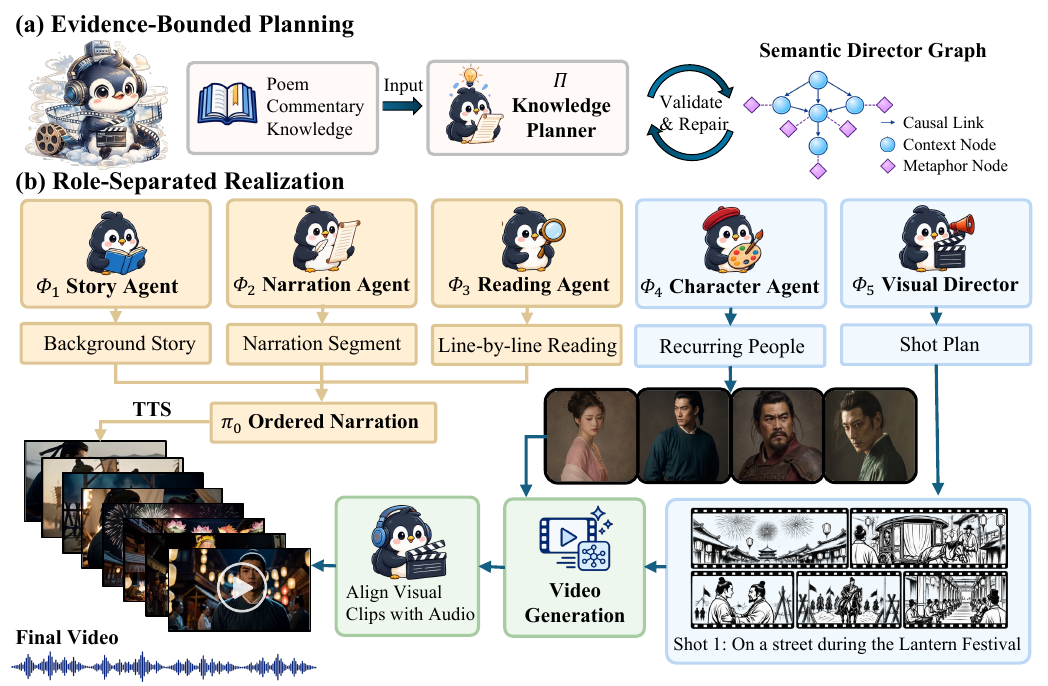}
\caption{The PoVi framework. A knowledge planner constructs and validates a
semantic director graph; five role-specialized stages realize the background
story, narration, close reading, character assets, and visual direction;
measured speech durations then control media generation, edit groups,
subtitles, and final assembly.}
\label{fig:pipeline}
\end{figure*}

\paragraph{Semantic Director Graph.}
The planner outputs
% \[
\begin{equation} 
G=(N,C,M,V),
\end{equation}
% \]
where $N$ contains contextual facts and events, $C\subseteq N\times N$ their causal relations, $M$ metaphorical mappings, and $V$ global visual constraints. Each factual or interpretive element $u\in N\cup C\cup M$ is grounded in evidence $\operatorname{cite}(u)\subseteq E$ and assigned a claim status $\sigma(u)$: \textsc{Plausible}, \textsc{Contested}, \textsc{Documented}, or \textsc{Consensus}.

\paragraph{Admissibility.}
The planner's output is accepted only if it passes structural, referential, and evidential
checks. Structural checks validate required fields and types; referential checks ensure valid references and coverage of all poems lines. Evidential checks prevent claims from exceeding the authority of their citations. Let $a^{*}(u)=\max_{e\in\mathrm{cite}(u)} a(e)$ for the strongest authority $u$
cites,
\begin{equation}
\begin{aligned}
\sigma(u)=\textsc{documented} &\Rightarrow a^{*}(u)\ge\textsc{med},\\
\sigma(u)=\textsc{consensus} &\Rightarrow a^{*}(u)=\textsc{high}.
\end{aligned}
\end{equation}
Citations must come from $E$, so the planner cannot invent a source. Claims supported only by model recall must remain \textsc{plausible} or \textsc{contested}.

\paragraph{Directed repair.}
Let $\mathcal{E}(G)$ list the checks a graph violates. Instead of resampling,
we iterate
% $$
\begin{equation} 
G^{(k+1)} \;=\; \Pi\bigl(x,\;G^{(k)},\;\mathcal{E}(G^{(k)})\bigr),
\qquad k<K,
\end{equation}
% $$
passing the model its own rejected output together with the specific
violations, and accept the first graph that passes all three. 

\begin{figure*}[t]
    \centering
    \includegraphics[width=\textwidth]{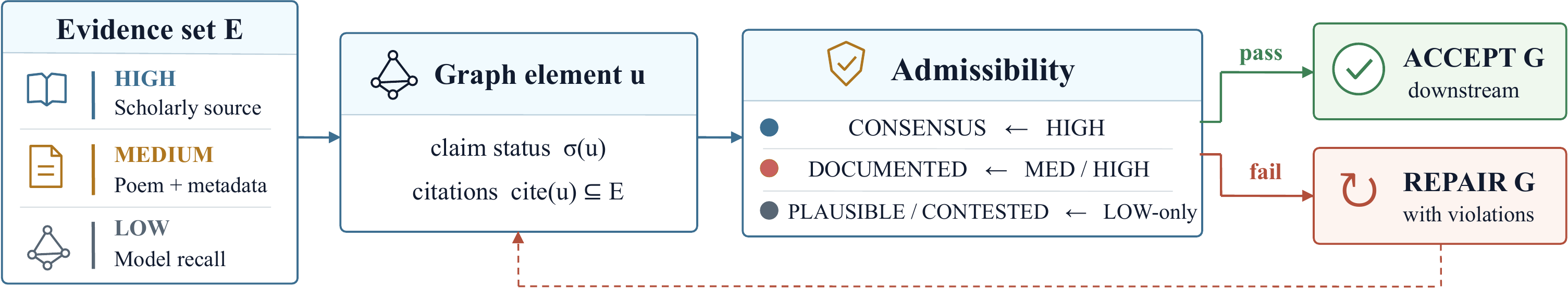}
    \caption{Evidence-bounded claim calibration. Each graph element is
    assigned a claim status and supporting citations. Evidence authority
    constrains the admissible status, while violations trigger directed
    repair before the graph is passed downstream.}
    \label{fig:correct}
\end{figure*}

Figure~\ref{fig:correct} summarizes how evidence authority
constrains the admissible status of each graph element and how violations
trigger directed repair.

\subsection{Role-Separated Realization}

Realization proceeds through five role-specialized agents, each with its own
instruction, output schema, and restricted view $\rho_i(G)$ of the semantic
director graph. The agents communicate only through validated intermediate
artifacts:
\begin{equation}
\begin{aligned}
s &= \phi_{1}(\rho_1(G)) &&\text{background story},\\
\tau &= \phi_{2}(s,\rho_2(G)) &&\text{narration segments},\\
r &= \phi_{3}(s,\rho_3(G)) &&\text{line-by-line reading}.
\end{aligned}
\end{equation}
Here, $\phi_1$ reconstructs the background story, $\phi_2$ segments it into
narration units and identifies the visual events to convey, and $\phi_3$
produces a line-by-line interpretation of the poem. 

The narration and reading
branches are then merged in a fixed programmatic order. The remaining stages enforce character consistency and convert the narration into a visual plan:
\begin{equation}
\begin{aligned}
\pi_0 &= \textsc{merge}(\tau,r) &&\text{ordered narration},\\
A &= \phi_{4}(\pi_0,\rho_4(G)) &&\text{recurring characters},\\
\pi &= \phi_{5}(\pi_0,A,\rho_5(G)) &&\text{shot descriptions}.
\end{aligned}
\end{equation}
Stage $\phi_4$ specifies the appearance of recurring characters, while
$\phi_5$ produces shot-level descriptions and groups them by narration
segment. All intermediate artifacts are validated
and repaired when necessary (Appendix~\ref{app:validation}).

\subsection{Audio-Clocked Assembly}

\begin{figure}[ht]
    \centering
    \includegraphics[width=\columnwidth]{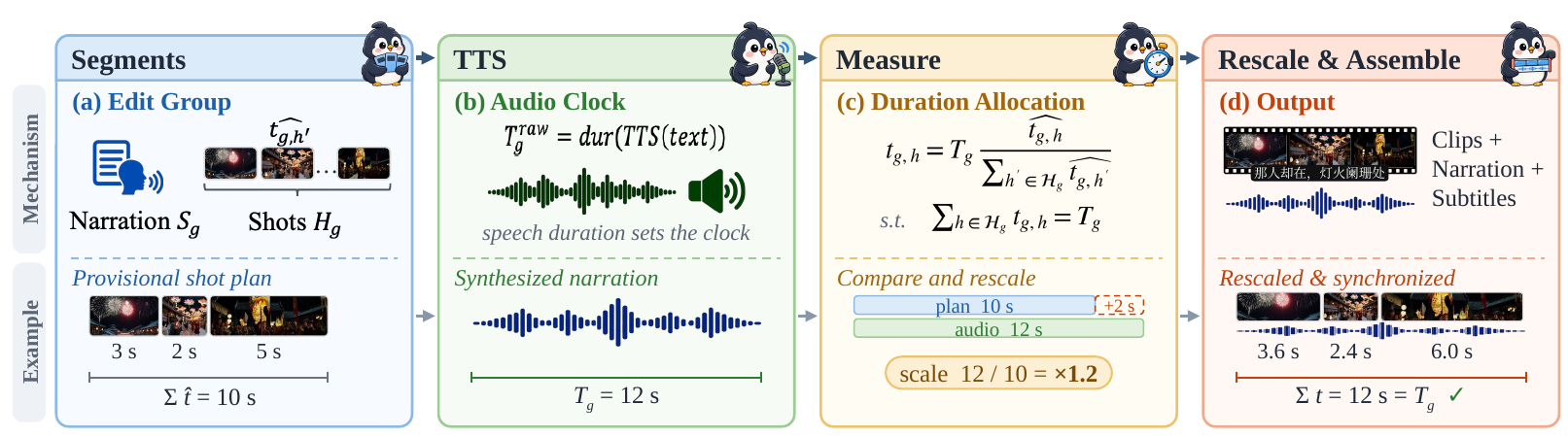}
    \caption{Audio-clocked assembly. Measured narration duration serves as the temporal reference for rescaling shot durations and final audiovisual assembly.}
    \label{fig:audio_clock}
\end{figure}

As illustrated in Figure~\ref{fig:audio_clock}, PoVi uses measured speech duration as the temporal reference for video assembly. Stage $\phi_2$ provides provisional segment durations, which are later rescaled after TTS to account for speaking-rate variation.

Stage $\phi_5$ groups narration segments and their corresponding shots into
\emph{edit groups}. Let $\mathcal{S}_g$ and $\mathcal{H}_g$ denote the ordered
narration segments and visual sequence of group $g$, respectively. Their
narration texts are concatenated and synthesized into a single audio track
with measured duration
% \[
\begin{equation} 
T_g^{\mathrm{raw}}
=
\operatorname{dur}\!\left(
\textsc{tts}\!\left(
\operatorname{concat}_{s\in\mathcal{S}_g}
\operatorname{text}(s)
\right)\right).
% \]
\end{equation}
Given the provisional duration $\hat{t}_{g,h}$ assigned by the visual director
to shot $h\in\mathcal{H}_g$, PoVi rescales the allocations to the measured
speech duration:
% \[
\begin{equation} 
t_{g,h}
=
T_g
\frac{\hat{t}_{g,h}}
{\sum_{h'\in\mathcal{H}_g}\hat{t}_{g,h'}},
\qquad
\sum_{h\in\mathcal{H}_g}t_{g,h}=T_g,
\end{equation}
% \]
where $T_g=T_g^{\mathrm{raw}}$ unless bounded speech-speed adjustment is needed
to fit the visual capacity.

For each shot, PoVi requests the shortest supported generation duration
covering $t_{g,h}$, then trims or mildly slows the resulting clip to match the
target. If an edit group remains infeasible, bounded speech acceleration is
applied; otherwise, the group is decomposed into finer units.

\section{Experiment}
\label{sec:exp}

\subsection{Setup}

% add citations
We evaluate all 120 gold-annotated works with $B=\emptyset$; gold
annotations are used only for evaluation. Gemini-3.5-Flash~\cite{googledeepmind2026gemini35flash},
DeepSeek-V4-Flash~\cite{deepseekai2026deepseekv4}, and Seed-2.0-Lite~\cite{seed2026seed2} are each compared with a single-call
baseline under the same backbone, video generator, and speech synthesizer.
For video, the Gemini planner is paired with MiniMax-H3~\cite{minimax-website}, Wan2.2-VACE~\cite{wan2025wan},
Veo-3.1~\cite{gallegos2025veo31}, and Seedance-2.0~\cite{seedance2026seedance}. Main tables report human ratings.
% ; automatic
% scores and human--Gemini agreement are reported in
% Appendix~\ref{app:full}. -> Section 4已经讲过了

%[TODO: 标注人数、背景、一致性]

\subsection{Narration}

Table~\ref{tab:text} reports the eight narration metrics. PoVi scores $0.11$ higher than the baseline on Gemini, $0.12$ on DeepSeek and $0.18$ on Seed-2.0-Lite. A single call has to recall the poet's history, choose what to say about it, and write the narration in one pass. Splitting those steps apart helps a weak model more than a strong one. The gain comes mostly from what the poem leaves out. S1 asks whether the narration covers the background a faithful explanation needs; it rises by $0.17$ to $0.23$ on every backbone. S4 asks why a metaphor's source licenses its target;
it rises by $0.06$ to $0.19$. Metrics that only need a close reading of the text hardly move: S6 checks that the narration is no more certain than the evidence allows, and it starts above $0.94$ everywhere and shifts by at most $0.02$.

D2 is the exception: it rises by \(0.17\) on Gemini and \(0.35\) on Seed, but falls by \(0.26\) on DeepSeek. D2 asks whether the prelude is built from concrete, storyboardable scenes and whether it leads naturally into the line-by-line explanation. Our framework writes this prelude as a continuous story before segmenting it for narration; this helps on two backbones but is less reliable on DeepSeek. 

\begin{table*}[t]
\centering
\small
\caption{Narration metrics, human ratings. D1 and D2 are 1--5 scores divided
by 5.}
\setlength{\tabcolsep}{3.2mm}

\begin{tabular}{llcccccccc}
\toprule
Backbone & System & S1 & S2 & S3 & S4 & S5 & S6 & D1 & D2 \\
\midrule

\multirow{2}{*}{Gemini-3.5-Flash}
& single call
& 0.611 & 0.931 & 0.851 & 0.617
& 0.889 & 0.943 & 0.815 & 0.633 \\

& ours
& \textbf{0.814}
& \best{0.951}
& \best{0.915}
& \best{0.789}
& \textbf{0.943}
& \best{0.965}
& \best{0.952}
& \textbf{0.800} \\

\midrule

\multirow{2}{*}{DeepSeek-V4-Flash}
& single call
& 0.533 & 0.756 & 0.512 & 0.334
& 0.688 & 0.955 & 0.725
& \best{0.825} \\

& ours
& \textbf{0.706}
& \textbf{0.894}
& \textbf{0.775}
& \textbf{0.521}
& \best{0.954}
& \textbf{0.957}
& \textbf{0.897}
& 0.569 \\

\midrule

\multirow{2}{*}{Seed-2.0-Lite}
& single call
& 0.625 & 0.478 & 0.400 & 0.365
& 0.665 & 0.944 & 0.906 & 0.334 \\

& ours
& \best{0.859}
& \textbf{0.692}
& \textbf{0.701}
& \textbf{0.428}
& \textbf{0.907}
& \textbf{0.961}
& \best{0.952}
& \textbf{0.688} \\

\bottomrule
\end{tabular}

\label{tab:text}
\end{table*}

\subsection{Video}

Table~\ref{tab:video} reports the video metrics across four generators. Our framework consistently improves or matches the single-call baseline,
with the largest average gain on Wan2.2-VACE ($+0.11$), suggesting that explicit semantic and shot-level planning is especially helpful for weaker generators.

\begin{table}[ht]
\centering
\small
\setlength{\tabcolsep}{5.4pt}
\caption{Video metrics, human ratings, Gemini planner. R3 and R5 are 1--5
scores divided by 5.}

\begin{tabular}{lccccc}
\toprule
& R1 & R2 & R3 & R4 & R5 \\
\midrule

\rowcolor{gray!15}
\multicolumn{6}{l}{\textit{MiniMax-H3}}\\
\quad single call
& 0.870
& 0.910
& 0.866
& \textbf{0.950}
& 0.882 \\

\quad ours
& \best{0.930}
& \textbf{0.930}
& \textbf{0.918}
& \textbf{0.950}
& \best{0.912} \\

\quad w/o ref.\ img.
& 0.812
& 0.796
& 0.840
& 0.000
& 0.870 \\

\addlinespace[2pt]

\rowcolor{gray!15}
\multicolumn{6}{l}{\textit{Wan2.2-VACE}}\\
\quad single call
& 0.671
& 0.788
& 0.698
& \textbf{0.900}
& 0.734 \\

\quad ours
& \textbf{0.832}
& \textbf{0.876}
& \textbf{0.898}
& \textbf{0.900}
& \textbf{0.824} \\

\quad w/o ref.\ img.
& 0.652
& 0.825
& 0.810
& 0.000
& 0.810 \\

\addlinespace[2pt]

\rowcolor{gray!15}
\multicolumn{6}{l}{\textit{Veo-3.1}}\\
\quad single call
& 0.699
& 0.980
& 0.897
& 0.950
& 0.811 \\

\quad ours
& \textbf{0.754}
& \best{0.998}
& \textbf{0.944}
& \best{0.984}
& \textbf{0.844} \\

\quad w/o ref.\ img.
& 0.534
& 0.982
& 0.880
& 0.000
& 0.830 \\

\addlinespace[2pt]

\rowcolor{gray!15}
\multicolumn{6}{l}{\textit{Seedance-2.0}}\\
\quad single call
& 0.770
& 0.958
& 0.939
& 0.880
& 0.896 \\

\quad ours
& \textbf{0.862}
& \textbf{0.978}
& \best{0.984}
& \textbf{0.978}
& \textbf{0.903} \\

\quad w/o ref.\ img.
& 0.765
& 0.974
& 0.810
& 0.000
& 0.880 \\

\bottomrule
\end{tabular}

\label{tab:video}
\end{table}

The largest gains on Wan2.2-VACE are in R1 ($0.671\!\rightarrow\!0.832$)
and R3 ($0.698\!\rightarrow\!0.898$), showing that planning mainly improves
visual requirement coverage and shot quality. Stronger generators gain less,
consistent with their better ability to follow explicit shot instructions
once the intended content is specified. Our framework gives the generator a reference portrait for each character who appears more than once. If we remove this stage and retain only the textual description, R4---which measures whether a character remains the same person across shots---drops from \(0.900\) to \(0.000\). R1 also decreases from \(0.832\) to \(0.652\), while the changes in R2, R3, and R5 are substantially smaller (\(-0.051\), \(-0.088\), and \(-0.014\), respectively). The automatic judge shows the same dominant collapse in R4.

\subsection{Cross-Modal Alignment}

Table~\ref{tab:av} reports the four cross-modal metrics under our framework. R7 asks whether each key event appears while it is being narrated, and it scores lower than R6 on both generators. D3 asks whether the picture adds something to what the narration is saying instead of just restating it. This is where the two generators differ most: $0.79$ on Wan2.2-VACE against $0.92$ on MiniMax-H3.

\begin{table}[ht]
\centering
\small
\setlength{\tabcolsep}{3.0mm}
\caption{Cross-modal metrics under our pipeline, human ratings. D3 and D4 are
1--5 scores divided by 5.}

\begin{tabular}{lcccc}
\toprule
Generator & R6 & R7 & D3 & D4 \\
\midrule

Wan2.2-VACE
& 0.881
& 0.812
& 0.785
& 0.878 \\

MiniMax-H3
& \best{0.904}
& \best{0.872}
& \best{0.919}
& \best{0.947} \\

\bottomrule
\end{tabular}

\label{tab:av}
\end{table}

\subsection{Comparison with Agentic Video Pipelines}

\begin{table*}[t]
\centering
\small
\setlength{\tabcolsep}{2.0mm}
\caption{Comparison with representative agentic video pipelines under the same DeepSeek-V4-Flash planning backbone, based on human
ratings. D1, D2 and R3 are 1--5 scores divided by 5, while the remaining are in
$[0,1]$.}

\begin{tabular}{lcccccccc@{\hskip 8pt}cccc}
\toprule
& \multicolumn{8}{c}{Narration}
& \multicolumn{4}{c}{Video} \\
\cmidrule(lr){2-9}
\cmidrule(lr){10-13}

System & S1 & S2 & S3 & S4 & S5 & S6 & D1 & D2
       & R1 & R2 & R3 & R4 \\
\midrule

ArcReel~\cite{arcreel}
& 0.780
& \best{1.000}
& \best{0.980}
& 0.900
& 0.800
& 0.840
& \best{1.000}
& 0.640
& 0.800
& 0.880
& 0.920
& 0.900 \\

Lumenx~\cite{lumenx}
& 0.700
& 0.940
& 0.780
& 0.680
& 0.960
& 0.900
& 0.980
& 0.680
& 0.780
& 0.880
& 0.880
& \best{1.000} \\

ours
& \best{0.880}
& 0.940
& 0.920
& \best{0.930}
& \best{0.970}
& \best{0.960}
& \best{1.000}
& \best{0.880}
& \best{0.880}
& \best{0.900}
& \best{0.960}
& \best{1.000} \\

\bottomrule
\end{tabular}

\label{tab:pipelines}
\end{table*}

We compare against ArcReel and Lumenx on a shared backbone
(Table~\ref{tab:pipelines}). Overall, PoVi achieves the best
or tied-best result on most metrics, with particularly clear advantages in
background coverage, discourse organization, and video realization. 

ArcReel is strongest where the poem can be read at face value: it reaches $1.00$ on naming a metaphor's source domain (S2) and $0.98$ on its target (S3). The gap opens on covering the background information the poem depends on (S1, $0.88$ vs.\ $0.78$ and
$0.70$) and on turning it into a prelude that leads into the reading (D2, $0.88$ vs.\ $0.64$ and $0.68$). This suggests that
the main advantage of PoVi lies not in recognizing isolated figurative elements,
but in reconstructing and organizing the implicit context needed for a coherent
explanation. Both baselines primarily direct content whose meaning is already assumed to be fixed by their input. PoVi explicitly inserts a semantic
reconstruction stage before directing, allowing background events and
interpretive structure to be established before they are converted into
narration and shots.

\begin{figure*}[ht]
\centering
\setlength{\tabcolsep}{3pt}
\renewcommand{\arraystretch}{0.9}

\newcommand{\csimage}[2]{%
  \begin{minipage}[b]{.228\textwidth}
    \centering
    \begin{tikzpicture}
      \node[
        anchor=south west,
        inner sep=0,
        outer sep=0
      ] (frame) at (0,0)
        {\includegraphics[width=\textwidth]{#1}};

      \node[
        anchor=south east,
        xshift=-2pt,
        yshift=2pt,
        fill=black,
        fill opacity=.60,
        text opacity=1,
        text=white,
        font=\scriptsize,
        inner xsep=3pt,
        inner ysep=1.5pt,
        outer sep=0
      ] at (frame.south east) {#2};
    \end{tikzpicture}
  \end{minipage}}

\newcommand{\csnote}[1]{%
  \parbox[t]{.228\textwidth}{%
    \raggedright\scriptsize #1}}
    
\hspace*{6pt}%
\includegraphics[width=\textwidth]{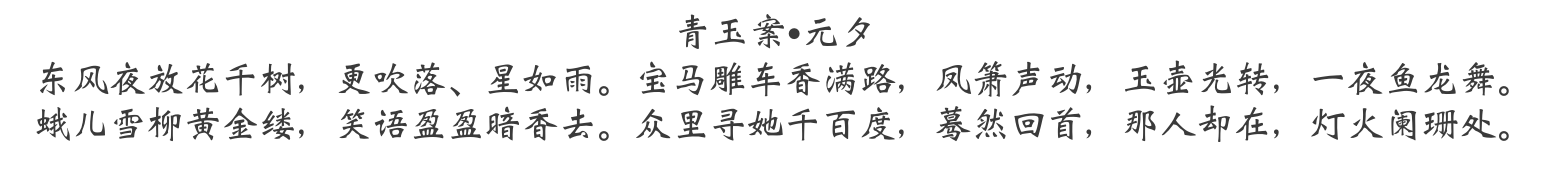}
\par\vspace{-4pt}

\begin{tabular*}{\textwidth}{%
  @{\extracolsep{\fill}}c@{\hspace{3pt}}cccc@{}}

\rotatebox{90}{\hspace{10pt}\small Single call}
& \csimage{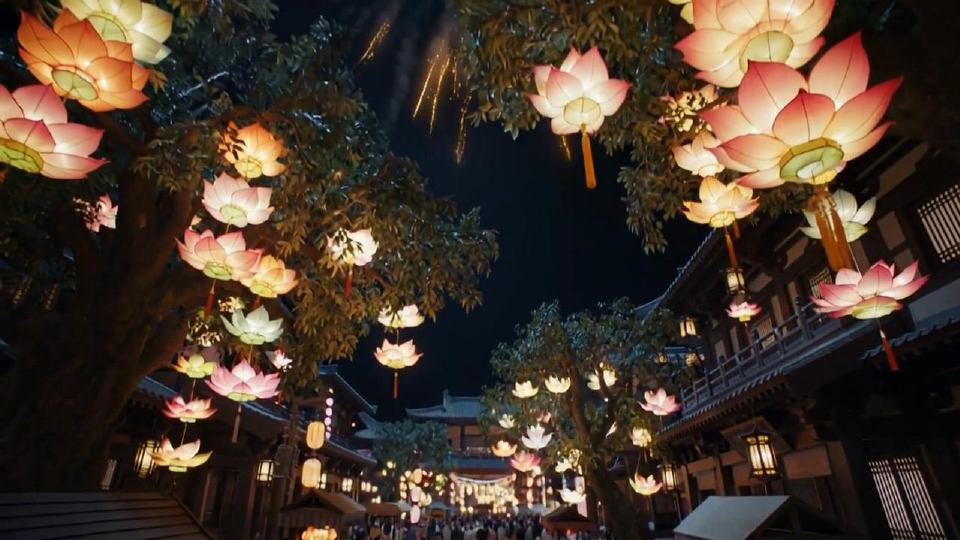}{00:04}
& \csimage{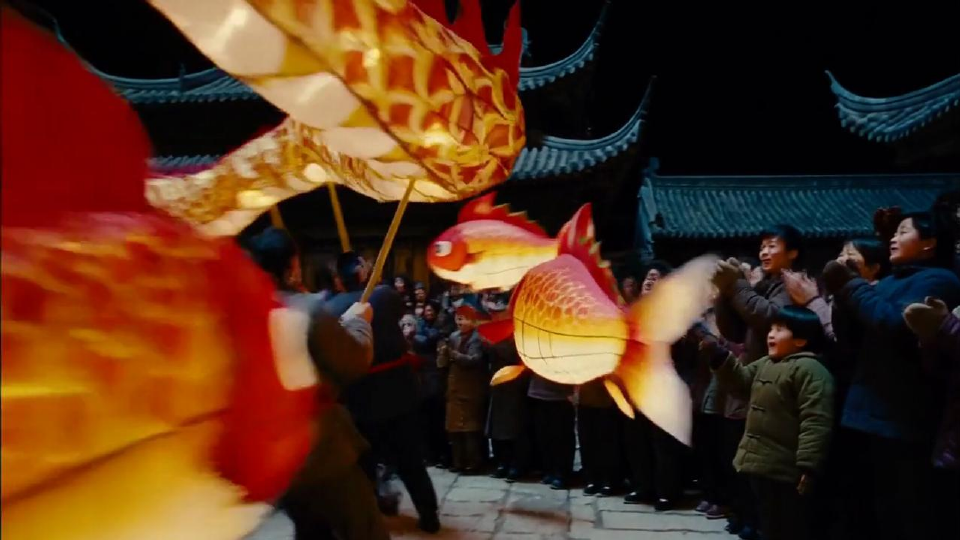}{00:22}
& \csimage{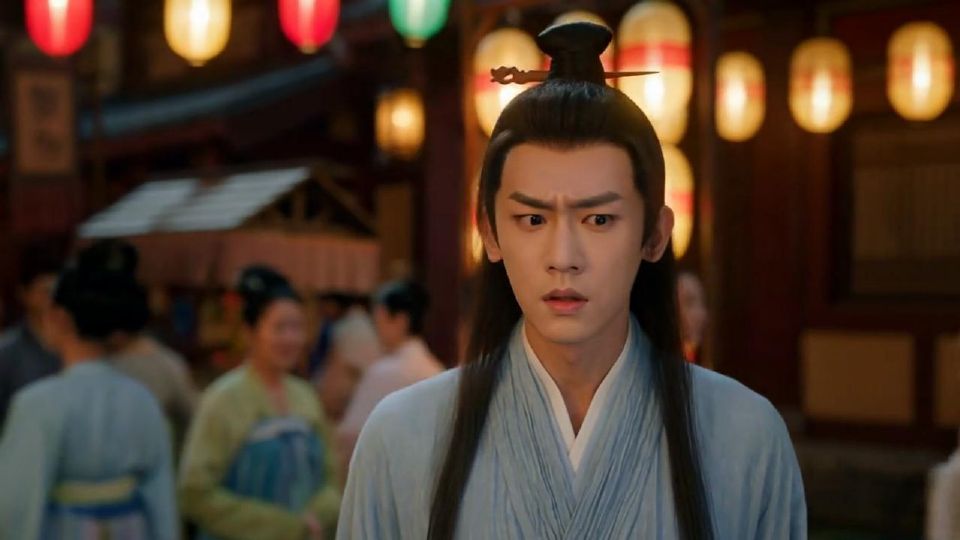}{00:40}
& \csimage{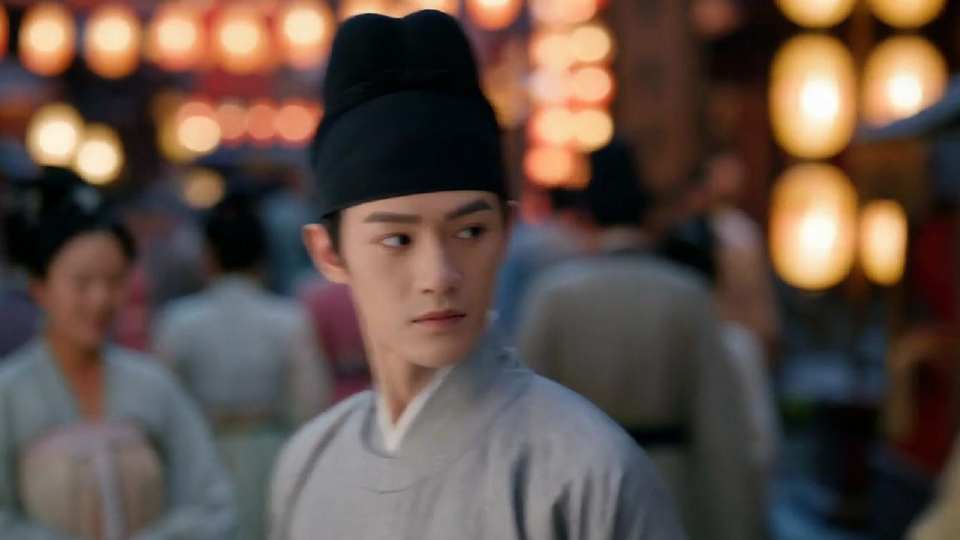}{00:46} \\[-4pt]

& \csnote{``The whole street blooms like a thousand trees of
  flowers.''}
& \csnote{``Fish and dragon lanterns dance among the joyous
  crowd.''}
& \csnote{``A scholar searches a thousand times for that one
  figure.''}
& \csnote{``He finds her in the loneliest corner.'' Inconsistent character appearance.} \\[8pt]
  
\rotatebox{90}{\hspace{22pt}\small PoVi}
& \csimage{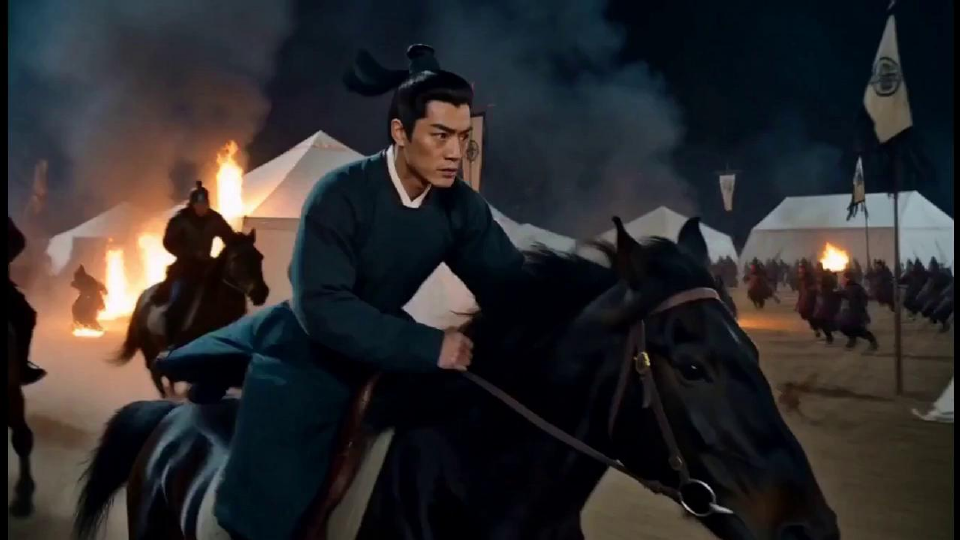}{00:44}
& \csimage{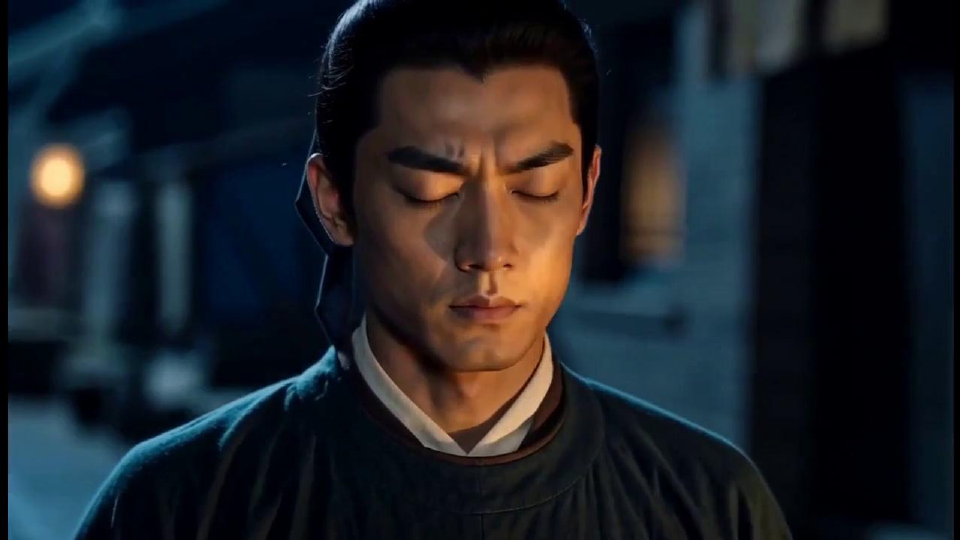}{01:58}
& \csimage{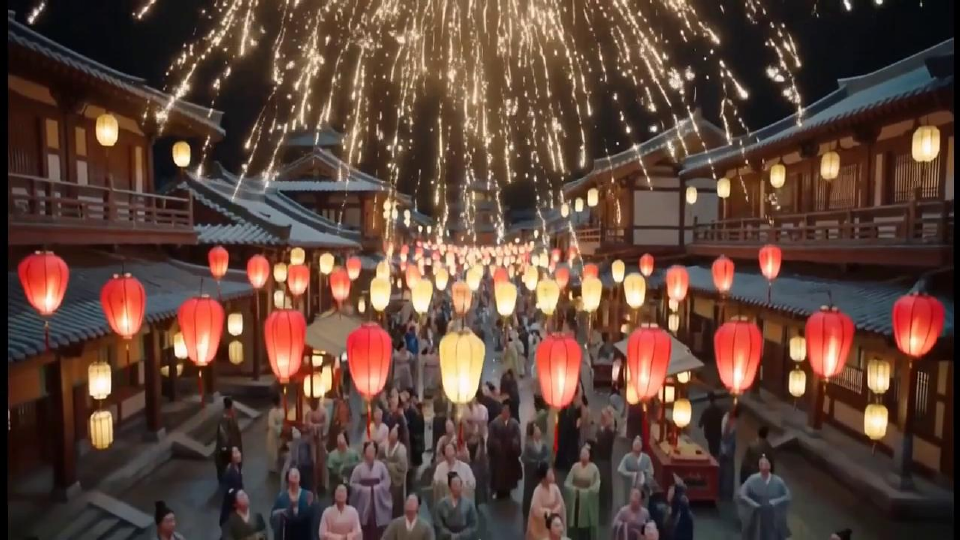}{02:58}
& \csimage{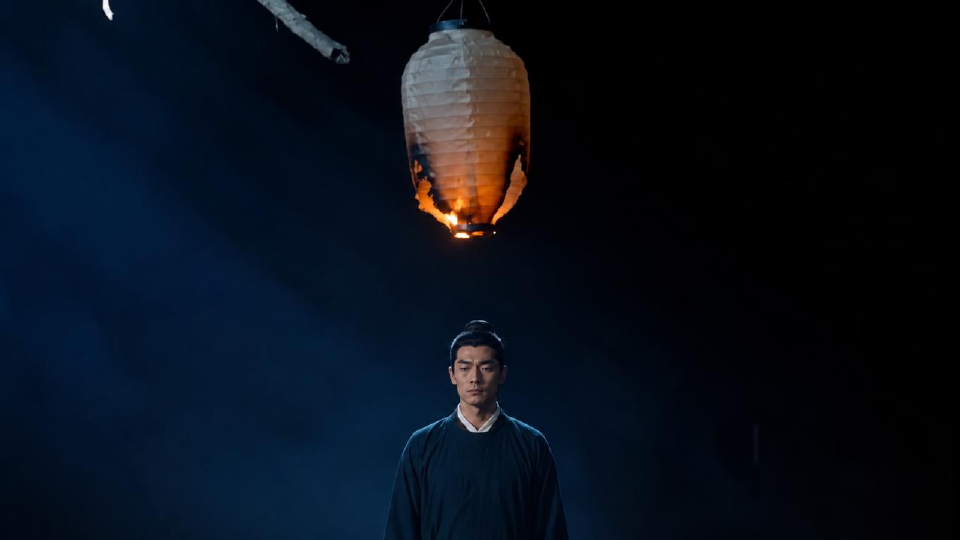}{04:22} \\[-4pt]

& \csnote{``He led only a few riders against tens of thousands of enemy troops, capturing the traitor alive.''}
& \csnote{``Even a whole city of dazzling splendour could not warm a heart set on recovering the lost land.''}
& \csnote{``He likens still lanterns to blooming spring flowers, and moving fireworks to a glittering rain of stars.''}
& \csnote{``\,`Yet there they stand, where the lanterns dim'---contrasting with splendour, suggesting aloofness.''}\\[4pt]

& \small (a) Historical context
& \small (b) Interpretive framing
& \small (c) Imagery and meaning
& \small (d) Closing reading \\
\end{tabular*}

\caption{Qualitative comparison of Xin Qiji's \emph{Qingyu An: Lantern Festival}. Both use Gemini-3.5-Flash with MiniMax-H3 (single call: 61\,s; PoVi: 4\,min 51\,s). Quotes translate the Chinese narration at each timestamp. Columns (c--d) show lyric readings; (a--b) show PoVi's historical prelude, absent from the baseline.}
\label{fig:case_study}
\end{figure*}

\subsection{Case Study}
\label{sec:case_study}

% Figure~\ref{fig:case_study} compares how both systems explain and
% visualize Xin Qiji's \emph{Qingyu An: Lantern Festival}. PoVi supplies
% biographical context that leads into its reading. The baseline opens
% directly with the festival; PoVi recounts the raid (a) and the court
% shelving his memorials, then contrasts the festival's splendour with
% his unfulfilled ambition to recover lost land (b).
%
% PoVi makes the figurative correspondences explicit, linking stationary
% lanterns to spring flowers and moving fireworks to a rain of stars (c).
% At the close, the baseline reports a meeting; PoVi reads the quiet
% corner against the splendour as reflecting the poet's aloofness (d).
% The background events and closing interpretation are specified in
% the director graph before generation.
%
% The closing shot also expresses this contrast visually. The baseline
% retains the festive crowd and bright lanterns; PoVi isolates the poet
% beneath a lantern against a dark background (d). Reference images keep
% him recognisable from 00:44 to 04:22, while the baseline scholar's
% topknot becomes a black cap between 00:40 and 00:46 (c--d), despite
% a prompt asking only for a head turn.

Figure~\ref{fig:case_study} compares how both systems explain and
visualize Xin Qiji's \emph{Qingyu An: Lantern Festival}. PoVi introduces
its reading through biographical context. The baseline opens with the
festival; PoVi recounts the raid (a) and the court shelving his memorials,
then contrasts festive splendour with his unfulfilled ambition
to recover lost land (b).

PoVi explicitly links stationary lanterns to spring flowers and moving
fireworks to a rain of stars (c). At the close, the baseline reports
a meeting while PoVi reads the quiet corner against the splendour as
reflecting the poet's aloofness (d). These background events and the
closing interpretation are specified in the director graph before generation
and carried into the visuals. The baseline
retains the festive crowd and bright lanterns; PoVi isolates the poet
beneath a lantern against a dark background (d). Reference images keep
him recognisable from 00:44 to 04:22; the baseline scholar's topknot
becomes a black cap between 00:40 and 00:46 (c--d), despite a prompt
requesting only a head turn.

\section{Conclusion}

\vspace{-2mm}
We presented PoVi, a benchmark and framework for poetry-to-video generation. Its evaluation protocol separately assesses semantic grounding, discourse coherence, and audiovisual realization against frozen annotations, preventing systems from explaining figurative meaning while visualizing only literal imagery. PoVi first reconstructs a poem's meaning, distinguishes model-recalled knowledge from documented evidence, and then converts this grounded interpretation into narration, character assets, shot-level plans, and temporally aligned audiovisual content. Across multiple planning backbones, PoVi consistently improves over single-call generation, with larger gains for weaker backbones. Video results show a similar trend: weaker generators benefit most from explicit semantic and shot-level planning, while stronger generators leave less room for improvement once the content is explicit. These findings show that stronger rendering alone cannot determine what a poem requires the video to communicate.

\section*{Limitations}

PoVi-Bench is deliberately balanced across Tang poetry, Song \emph{ci}, and three difficulty levels. Nevertheless, its 120 works cannot fully capture the authorial, historical, and formal diversity of classical Chinese poetry. Automatic evaluation by Gemini-3.5-Flash shows moderate agreement with human ratings but saturates on some visual metrics. We therefore use human ratings as the reference. Finally, PoVi requires eight model calls per poem before repairs, compared with one call for the baseline, increasing computational cost.


\bibliography{ref}

\clearpage
\appendix

\section{Evaluation Details}
\label{app:metrics}

This section provides the full definitions and scoring procedures for
the 17 PoVi metrics summarized in Figure~\ref{fig:metrics}. All gold
annotations are frozen before evaluation. Each metric is evaluated
independently against the corresponding gold information. For
video-based metrics, evidence is first collected over temporal windows
and then aggregated so that each gold item is adjudicated once using
all relevant evidence from the full video.

\subsection{Metric Definitions}

Table~\ref{tab:metrics} provides the full definitions, scoring scales, and associated gold fields for all PoVi evaluation metrics.
%gemini table

\begin{table*}[!t]
\centering\small
\caption{The 17 primary metrics of the benchmark, grouped by the three challenges of the task.}
\setlength{\tabcolsep}{3mm}
% \scriptsize
% \setlength{\tabcolsep}{3pt}
% \renewcommand{\arraystretch}{0.88}
\begin{tabular}{@{}llcp{6.6cm}l@{}}
\toprule
\textbf{ID} & \textbf{Metric} & \textbf{Scale} & \textbf{What it asks} & \textbf{Gold field} \\
\midrule
\multicolumn{5}{@{}l}{\textit{Semantic grounding --- is the poem understood?}}\\
S1 & Information Coverage        & $[0,1]$ & Is the information a faithful explanation must convey actually conveyed? & \texttt{required\_info} \\
S2 & Metaphor Source             & $[0,1]$ & Is the vehicle imagery explicitly identified? & \texttt{metaphor\_maps} \\
S3 & Metaphor Target             & $[0,1]$ & Is the abstract referent stated without over-generalization? & \texttt{metaphor\_maps} \\
S4 & Mapping Bridge              & $[0,1]$ & Is it explained \emph{why} the source licenses the target? & \texttt{metaphor\_maps} \\
S5 & Metaphor Contextual Support & $[0,1]$ & Is the figurative reading supported by text, context or tradition rather than free association? & \texttt{metaphor\_maps} \\
S6 & Interpretive Calibration    & $[0,1]$ & Does the epistemic stance match the annotated status of the reading? & \texttt{status} \\
\midrule
\multicolumn{5}{@{}l}{\textit{Discourse coherence --- is the understanding organized into an explanation?}}\\
D1 & Narrative Structure         & $1$--$5$ & Are event order, state change, causality and scene transitions correct? & \texttt{key\_events} \\
D2 & Prelude--Explication Linkage & $1$--$5$ & Is the background dramatized concretely and does it lead into line-by-line explication? & \texttt{key\_events} \\
D3 & Modality Complementarity    & $1$--$5$ & Does the visual channel add information rather than duplicate the narration? & --- \\
D4 & Audiovisual Pacing          & $1$--$5$ & Are shot duration, transition frequency and information density coherent across phases? & --- \\
\midrule
\multicolumn{5}{@{}l}{\textit{Audiovisual realization --- is the explanation realized and aligned?}}\\
R1 & Visual Requirement Coverage & $[0,1]$ & Do the required visual elements appear at all? & \texttt{visual\_req} \\
R2 & Visual Semantic Correctness & $[0,1]$ & Is what \emph{is} shown correct in identity, period, space, action and outcome? & \texttt{visual\_req} \\
R3 & Static Visual Quality       & $1$--$5$ & Sharpness, exposure, colour, texture, subject legibility. & --- \\
R4 & Character Identity Continuity & $[0,1]$ & Is the same narrative character still the same person across shots? & --- \\
R5 & Visual Metaphor Realization & $1$--$5$ & Are source, bridge and target realized visually? & \texttt{metaphor\_maps} \\
R6 & Cross-modal Correspondence  & $[0,1]$ & Do concurrent narration and image refer to the same content? & --- \\
R7 & Narration--Event Synchrony  & $[0,1]$ & Does each key event appear when it is narrated? & \texttt{key\_events} \\
\bottomrule
\end{tabular}
\label{tab:metrics}
\end{table*}

\subsection{Scoring and Aggregation}
Coverage-based metrics are evaluated against frozen gold items.
For S1, each required item is scored as fully covered, partially
covered, or absent, and aggregated using its annotation weight:
\[
\mathrm{S1}=\frac{\sum_i w_i c_i}{\sum_i w_i},
\qquad c_i\in\{1,0.5,0\}.
\]
where $w_i$ is the importance weight of gold item $i$.
R1 uses the same formulation over visual requirements.

R1 and R2 are evaluated separately. R1 measures whether a required
visual element appears, whereas R2 evaluates the semantic correctness
of what is actually shown. Missing content is therefore not penalized
again as an incorrect realization.

For duration-dependent visual quality, R3 is aggregated over temporal
windows:
\[
\mathrm{R3}=\frac{\sum_w d_w s_w}{\sum_w d_w}.
\]

where $s_w$ is the quality score of window $w$ and $d_w$ is its
duration.

R6 evaluates the relation between temporally concurrent narration and
visual content. We assign $1$ to aligned or complementary content,
$0.75$ to redundant content, $0.25$ to unrelated content, and $0$ to
conflicting content.

Metaphor understanding is evaluated at multiple levels. S2 and S3
measure whether the narration identifies the metaphor source and
target, while S4 evaluates whether it explains the bridge connecting
them. S5 further evaluates whether the interpretation is supported by
the poem, historical context, or established interpretive evidence.

\paragraph{Cross-modal metaphor evaluation.}
Metaphor is evaluated independently in narration and video:
S2--S4 assess whether the narration identifies and explains the
source--target relation, whereas R5 assesses whether the same relation
is realized visually.

For metrics defined on a $1$--$5$ scale, the original score is retained
for human evaluation and divided by $5$ when reported together with
metrics whose native range is $[0,1]$.

\section{Automatic Evaluation}
\label{app:full}

Human ratings are used as the reference evaluation throughout the main paper. To provide a lower-cost alternative for larger-scale evaluation, we additionally score the same outputs with Gemini-3.5-Flash using the same metric definitions and frozen gold annotations. Scores originally
defined on the $1$--$5$ scale are divided by $5$ in the tables below.

\subsection{Judge Scores}
%\label{app:full}

Tables~\ref{tab:app-text}--\ref{tab:app-pipe} report the
Gemini-3.5-Flash scores for the experimental conditions in
Section~\ref{sec:exp}.

\begin{table*}[t]
\centering
\small
\setlength{\tabcolsep}{3.2mm}
\caption{Narration metrics, Gemini-3.5-Flash judge. Compare with human ratings
in Table~\ref{tab:text}.}

\begin{tabular}{llcccccccc}
\toprule
Backbone & System & S1 & S2 & S3 & S4 & S5 & S6 & D1 & D2 \\
\midrule

\multirow{2}{*}{Gemini-3.5-Flash}
& single call
& 0.729
& 0.976
& 0.929
& 0.670
& 0.941
& 0.943
& 0.978
& 0.823 \\

& ours
& \best{0.914}
& \best{0.983}
& \best{0.981}
& \best{0.953}
& \best{0.980}
& \best{0.979}
& \textbf{0.981}
& \best{0.954} \\

\midrule

\multirow{2}{*}{DeepSeek-V4-Flash}
& single call
& 0.634
& 0.960
& 0.863
& 0.816
& 0.898
& \textbf{0.966}
& 0.933
& 0.806 \\

& ours
& \textbf{0.906}
& \textbf{0.976}
& \textbf{0.954}
& \textbf{0.934}
& \textbf{0.952}
& 0.914
& \best{0.983}
& \textbf{0.921} \\

\midrule

\multirow{2}{*}{Seed-2.0-Lite}
& single call
& 0.464
& 0.523
& 0.500
& 0.202
& 0.694
& 0.823
& 0.912
& 0.370 \\

& ours
& \textbf{0.821}
& \textbf{0.758}
& \textbf{0.724}
& \textbf{0.524}
& \textbf{0.922}
& \textbf{0.965}
& \textbf{0.969}
& \textbf{0.651} \\

\bottomrule
\end{tabular}

\label{tab:app-text}
\end{table*}

\begin{table}[t]
\centering
\small
\setlength{\tabcolsep}{1.9mm}
\caption{Video metrics, Gemini-3.5-Flash judge. R3 and R5 are 1--5 scores
divided by 5. Compare with human ratings in Table~\ref{tab:video}.}

\begin{tabular}{lccccc}
\toprule
& R1 & R2 & R3 & R4 & R5 \\
\midrule

\multicolumn{6}{l}{\textit{MiniMax-H3}}\\
\quad single call
& 0.811
& \best{1.000}
& 0.985
& \best{1.000}
& 0.821 \\

\quad ours
& \best{0.831}
& \best{1.000}
& \textbf{0.988}
& \best{1.000}
& \textbf{0.874} \\

\addlinespace[2pt]

\multicolumn{6}{l}{\textit{Wan2.2-VACE}}\\
\quad single call
& 0.682
& 0.880
& \textbf{0.938}
& \best{1.000}
& 0.800 \\

\quad ours
& \textbf{0.790}
& \textbf{0.881}
& 0.935
& \best{1.000}
& \textbf{0.848} \\

\quad w/o ref.\ img.
& 0.689
& 0.813
& 0.908
& 0.000
& 0.807 \\

\addlinespace[2pt]

\multicolumn{6}{l}{\textit{Veo-3.1}}\\
\quad single call
& 0.731
& 0.998
& 0.973
& \best{1.000}
& 0.830 \\

\quad ours
& \textbf{0.779}
& \best{1.000}
& \textbf{0.999}
& \best{1.000}
& \textbf{0.863} \\

\addlinespace[2pt]

\multicolumn{6}{l}{\textit{Seedance-2.0}}\\
\quad single call
& 0.820
& \best{1.000}
& 0.995
& \best{1.000}
& \best{0.888} \\

\quad ours
& \textbf{0.824}
& \best{1.000}
& \best{1.000}
& \best{1.000}
& 0.886 \\

\bottomrule
\end{tabular}

\label{tab:app-video}
\end{table}

\begin{table}[t]
\centering
\small
\setlength{\tabcolsep}{3.0mm}
\caption{Cross-modal metrics, Gemini-3.5-Flash judge. D3 and D4 are 1--5 scores
divided by 5. Compare with human ratings in Table~\ref{tab:av}.}

\begin{tabular}{lcccc}
\toprule
Generator & R6 & R7 & D3 & D4 \\
\midrule

Wan2.2-VACE
& 0.890
& 0.817
& 0.937
& 0.902 \\

MiniMax-H3
& \best{0.940}
& \best{0.925}
& 0.968
& 0.975 \\

Seedance-2.0
& 0.933
& 0.897
& \best{0.992}
& \best{0.986} \\

\bottomrule
\end{tabular}

\label{tab:app-av}
\end{table}

\begin{table*}[t]
\centering
\small
\setlength{\tabcolsep}{2.5mm}
\caption{Comparison with agentic video pipelines, Gemini-3.5-Flash judge.
Compare with human ratings in Table~\ref{tab:pipelines}.}

\begin{tabular}{lcccccccc@{\hskip 8pt}cccc}
\toprule
& \multicolumn{8}{c}{Narration}
& \multicolumn{4}{c}{Video} \\
\cmidrule(lr){2-9}
\cmidrule(lr){10-13}

System & S1 & S2 & S3 & S4 & S5 & S6 & D1 & D2
       & R1 & R2 & R3 & R4 \\
\midrule

ArcReel
& 0.830
& \best{0.980}
& 0.934
& \best{0.972}
& 0.944
& 0.800
& \best{1.000}
& 0.560
& 0.767
& \best{1.000}
& 0.988
& 0.900 \\

Lumenx
& 0.818
& 0.934
& 0.894
& 0.760
& 0.934
& 0.860
& \best{1.000}
& 0.440
& 0.702
& \best{1.000}
& \best{1.000}
& \best{1.000} \\

Ours
& \best{0.906}
& 0.976
& \best{0.954}
& 0.934
& \best{0.952}
& \best{0.914}
& 0.983
& \best{0.921}
& \best{0.839}
& \best{1.000}
& \best{1.000}
& \best{1.000} \\

\bottomrule
\end{tabular}

\label{tab:app-pipe}
\end{table*}

\subsection{Agreement with Human Ratings}
We evaluate the reliability of Gemini-3.5-Flash by comparing its
automatic scores with human ratings on the eight narration metrics.
At the atomic level, each data point corresponds to a matched
system--poem--metric tuple. We report Spearman rank correlation (SRCC),
Kendall rank correlation (KRCC), Pearson correlation (PLCC), and
normalized root mean squared error (nRMSE).

\begin{table}[t]
\centering
\small
\setlength{\tabcolsep}{4pt} 
\caption{Agreement between Gemini-3.5-Flash and human ratings on the eight narration metrics over all 120 works in PoVi-Bench.}
\begin{tabular}{@{}lcccc@{}}
\toprule
Evaluator & SRCC $\uparrow$ & KRCC $\uparrow$ & PLCC $\uparrow$ & nRMSE $\downarrow$ \\
\midrule
Gemini-3.5-Flash & 0.505 & 0.429 & 0.503 & 0.162 \\
\bottomrule
\end{tabular}
\label{tab:meta}
\end{table}

The positive rank and linear correlations indicate that the automatic evaluator broadly tracks human judgments, while the remaining gap motivates our use of human ratings as the primary evaluation throughout the paper.

\section{Validation and Repair}
\label{app:validation}

Every model-generated intermediate artifact is validated by program before it is passed to the next stage. This section describes the validation rules, locked fields, downstream checks, and repair budget used by the PoVi framework.

\subsection{Graph Validation}

The semantic director graph is validated at three levels. If any check fails, the rejected graph and the corresponding violations are returned to the planner for repair.

\paragraph{Schema.}
We verify field types, required fields, and the absence of unexpected properties. JSON Schema meta-fields such as
\texttt{\$schema}, \texttt{\$id}, \texttt{title}, and \texttt{type} are rejected when copied into the generated instance.

\paragraph{Rules.}
Identifiers must follow the required format; every cross-reference must resolve to a declared node; every poem line must be covered by at least one node; causal links may refer only to declared context nodes; and the causal graph must remain acyclic.

\paragraph{Grounding.}
We additionally verify whether each cited source licenses the claimed
epistemic status:
\begin{itemize}[leftmargin=1.2em,itemsep=1pt,topsep=3pt,parsep=0pt]
\item \textsc{documented} requires at least one medium- or
high-authority source.
\item Model-recalled evidence cannot license \textsc{documented}.
\item \textsc{consensus} requires a high-authority source.
\item Model-recalled evidence cannot license \textsc{consensus}.
\item Every citation must resolve to an evidence identifier already
declared in the evidence set.
\end{itemize}

Claims supported only by model-recalled knowledge must therefore remain
\textsc{plausible} or \textsc{contested}. The planner cannot introduce
new evidence identifiers during graph generation or repair.

\subsection{Locked Fields}

Several fields are fixed before the planner is called and restored
after generation so that the planner cannot silently modify the task
input. These include the schema version, document identifier, poem text
and line identifiers, task profile, provided evidence sources and the minimum poem-card duration.

Normalization also removes fields that the planner is not authorized
to produce, including background-story prose, teaching scripts,
character assets, and shot descriptions. These fields belong to the
downstream realization stages and are therefore discarded if they
appear prematurely.

\subsection{Downstream Checks}

Each realization stage has its own deterministic validation rules.
Quoted poem text must match the source exactly, character for character.
A narration segment contains two to five visual events, with a minimum
of
\[
\left\lceil \frac{\text{seconds}}{8} \right\rceil
\]
events so that a single generated clip is not required to cover a
substantially longer interval.

Narration may not refer to its own production process. Provisional
segment durations proposed by the model are replaced by
character-count-based estimates and clamped to $[3,45]$ seconds. These
values are used only for planning; the final timeline is determined
from the measured synthesized speech duration as described in
Section~\ref{sec:method}.

\subsection{Retry Budget}

Table~\ref{tab:retries} summarizes the model-call and repair budget.
Stage $\phi_3$ uses two model calls---a close-reading planner followed
by a close-reading realizer---while functioning as a single
role-specialized stage in the framework.

\begin{table}[h]
\centering
\small
\setlength{\tabcolsep}{5.5mm} 
\caption{Model calls per poem. A run with no repairs uses eight calls; the worked
example used twelve.}
\begin{tabular}{@{}lcc@{}}
\toprule
Stage & First call & Max repairs \\
\midrule
Knowledge inventory & 1 & 0 \\
Graph generation & 1 & 3 \\
Context story & 1 & 3 \\
Story narration & 1 & 3 \\
Close-reading planner & 1 & 2 \\
Close-reading realizer & 1 & 2 \\
Character assets & 1 & 2 \\
Visual direction & 1 & 2 \\
\midrule
Total & 8 & 17 \\
\bottomrule
\end{tabular}
\label{tab:retries}
\end{table}

There is no critic model. Rejection is decided by program, and the same stage that produced an invalid output is asked to repair it. If a stage exhausts its repair budget, the run is marked as failed rather than falling back to a degraded output. No run in the reported experiments exhausted its repair budget.

\section{Cases Analysis}
\label{seccase}

We present three illustrative high-scoring examples covering Tang poetry,
Song ci, and different levels of textual complexity. For each example, we
select nine frames from the generated video. Frames (a)--(f) visualize the
background story and its transition to the poem, while frames (g)--(i)
illustrate the line-by-line reading. These examples are qualitative
illustrations rather than representative samples of the complete evaluation
set. Gemini scores evaluate the final narration only and do not measure visual
quality.

\subsection{\emph{Jinse}: Challenging Tang-Poetry Example}
\label{app:case_jinse}

\begin{figure*}[!t]
\centering
\setlength{\tabcolsep}{2pt}
\renewcommand{\arraystretch}{0.85}
\begin{tabular}{ccc}
\includegraphics[width=.315\textwidth]{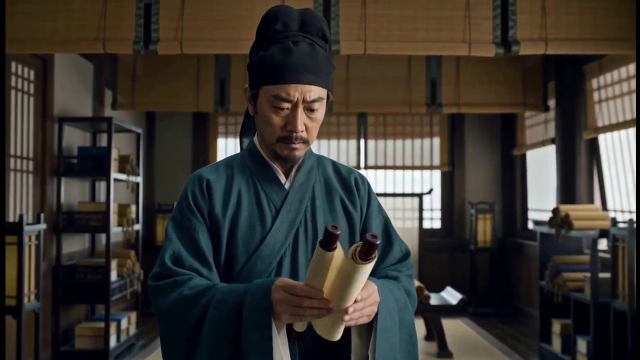} &
\includegraphics[width=.315\textwidth]{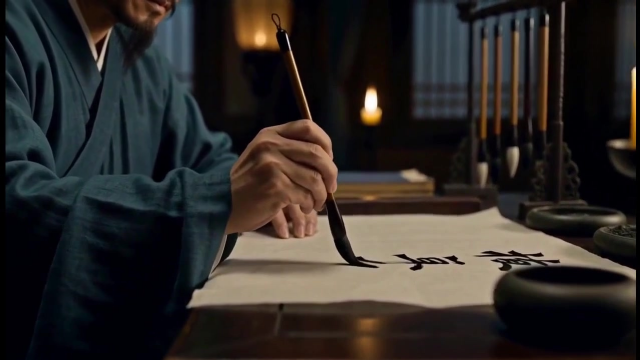} &
\includegraphics[width=.315\textwidth]{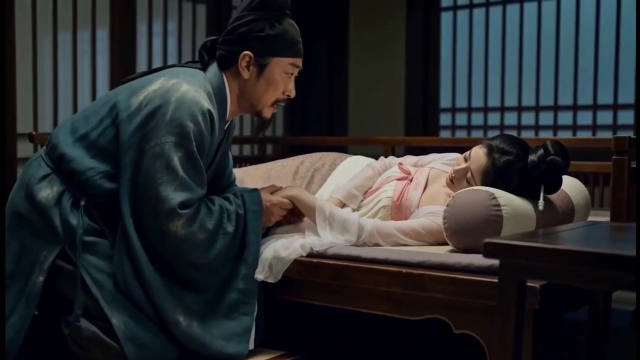} \\
\scriptsize (a) Official correspondence &
\scriptsize (b) Scholarly writing &
\scriptsize (c) Personal loss \\[3pt]
\includegraphics[width=.315\textwidth]{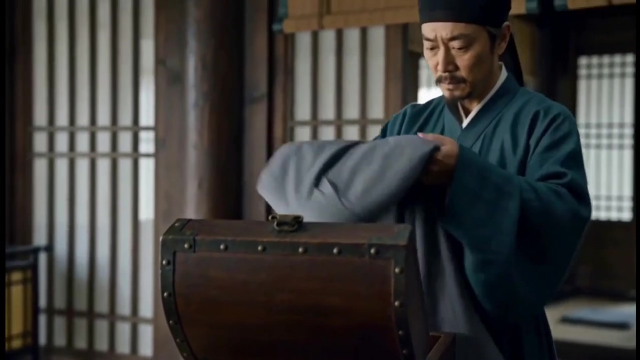} &
\includegraphics[width=.315\textwidth]{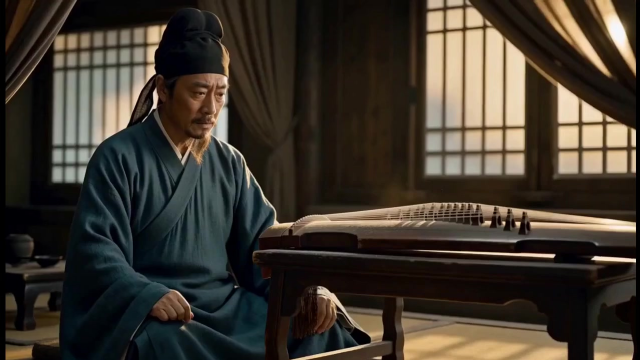} &
\includegraphics[width=.315\textwidth]{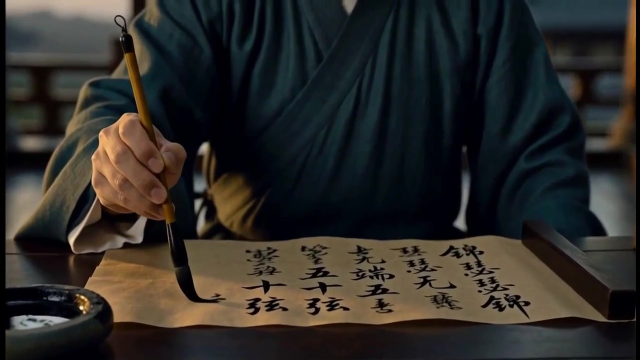} \\
\scriptsize (d) Departure for Zizhou &
\scriptsize (e) The zither motif &
\scriptsize (f) Transition to the poem \\[3pt]
\includegraphics[width=.315\textwidth]{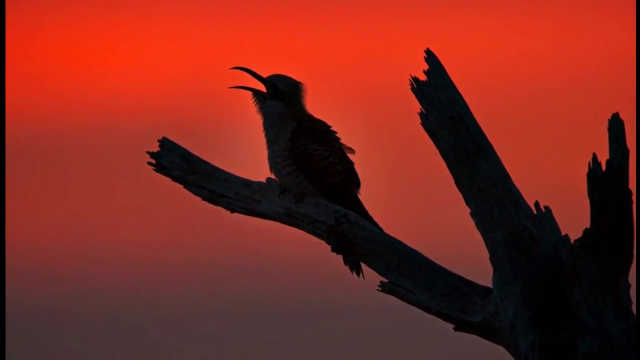} &
\includegraphics[width=.315\textwidth]{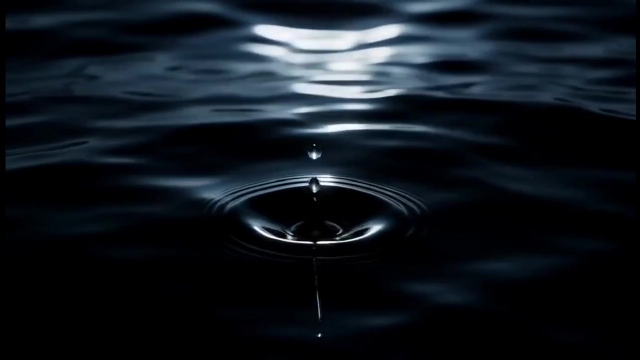} &
\includegraphics[width=.315\textwidth]{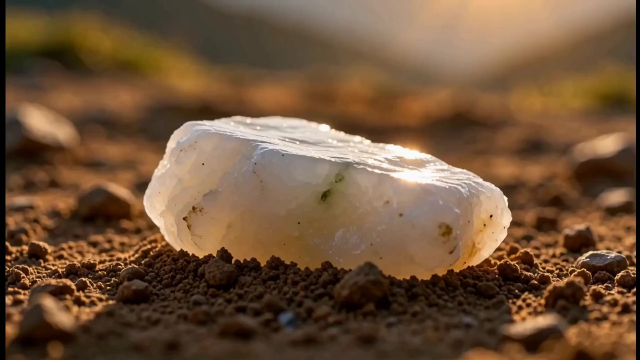} \\
\scriptsize (g) The cuckoo allusion &
\scriptsize (h) Pearl-tear imagery &
\scriptsize (i) Lantian jade
\end{tabular}
\caption{Selected frames from the generated teaching video for Li Shangyin's
\emph{Jinse}. Frames (a)--(f) visualize the background story and its
transition to the poem; frames (g)--(i) visualize allusive imagery used in
the line-by-line reading.}
\label{fig:case_jinse}
\end{figure*}

\noindent\textbf{Gemini-based evaluation.}
The narration receives 1.00 for Required Information Coverage and all five
metaphor criteria (Source Identification, Target Identification, Mapping
Justification, Contextual Grounding, and Interpretive Calibration), together
with 5/5 for both Narrative Coherence and Storytelling--Teaching Transition.

\paragraph{Analysis.}
The generated narration adopts one possible interpretive pathway, connecting
the poem with political frustration, personal loss, and displacement. Because
the biographical meaning of \emph{Jinse} remains contested, these connections
should be understood as an instructional framing rather than as definitive
claims about the poem's occasion or intended meaning. The zither provides a
recurring visual anchor between the background story and the poem. The later
frames render several allusive images as distinct visual units, helping the
viewer follow the poem while retaining its interpretive ambiguity.

% \clearpage

\subsection{\emph{Mooring by Maple Bridge}: Introductory
Tang-Poetry Example}
\label{app:case_fengqiao}

\begin{figure*}[!t]
\centering
\setlength{\tabcolsep}{2pt}
\renewcommand{\arraystretch}{0.85}
\begin{tabular}{ccc}
\includegraphics[width=.315\textwidth]{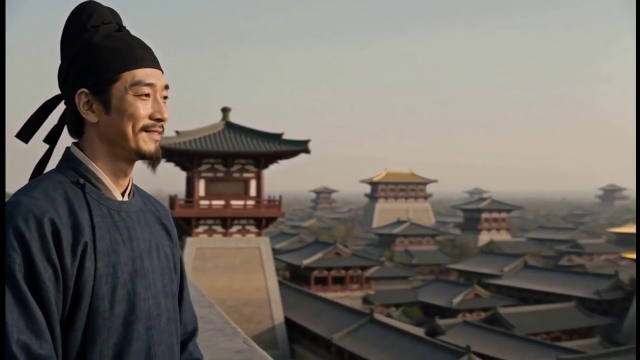} &
\includegraphics[width=.315\textwidth]{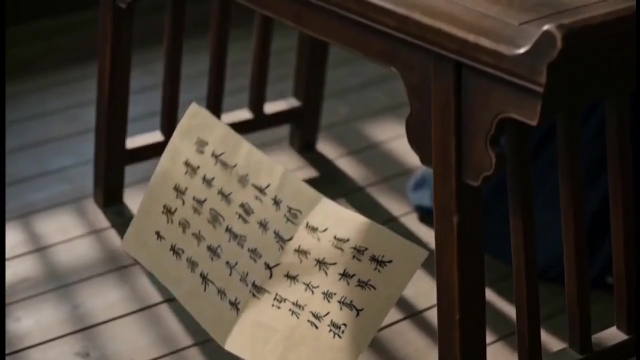} &
\includegraphics[width=.315\textwidth]{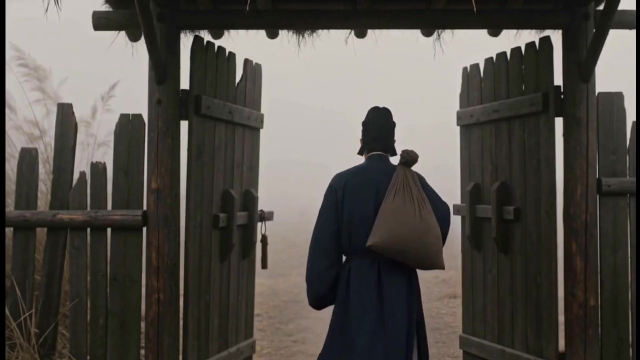} \\
\scriptsize (a) Examination success &
\scriptsize (b) News of unrest &
\scriptsize (c) Southward displacement \\[3pt]
\includegraphics[width=.315\textwidth]{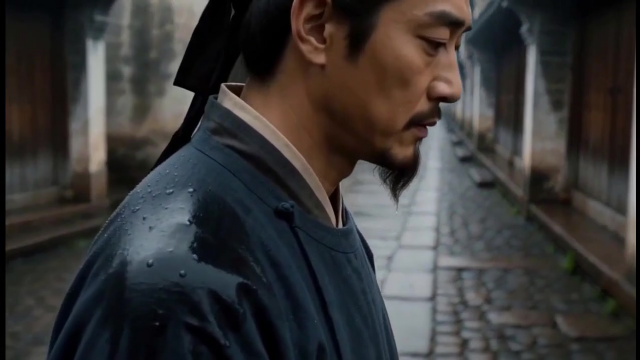} &
\includegraphics[width=.315\textwidth]{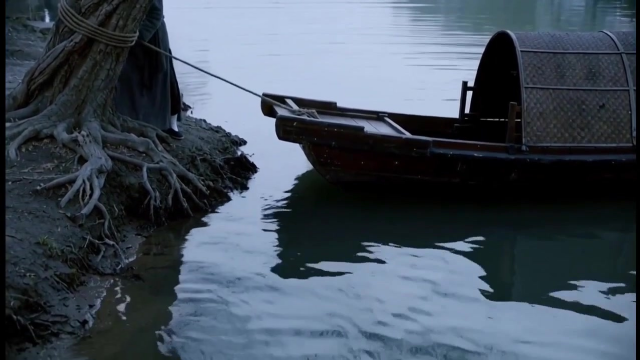} &
\includegraphics[width=.315\textwidth]{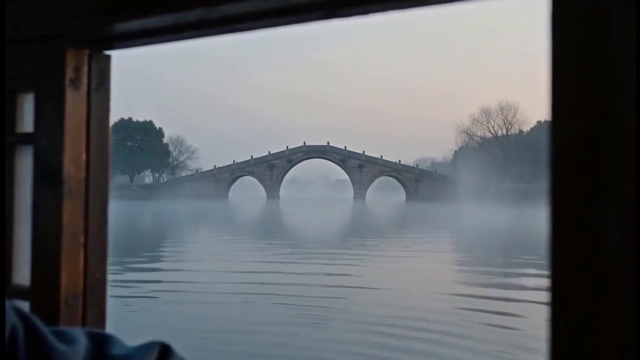} \\
\scriptsize (d) Life in exile &
\scriptsize (e) Mooring at night &
\scriptsize (f) Arrival at Maple Bridge \\[3pt]
\includegraphics[width=.315\textwidth]{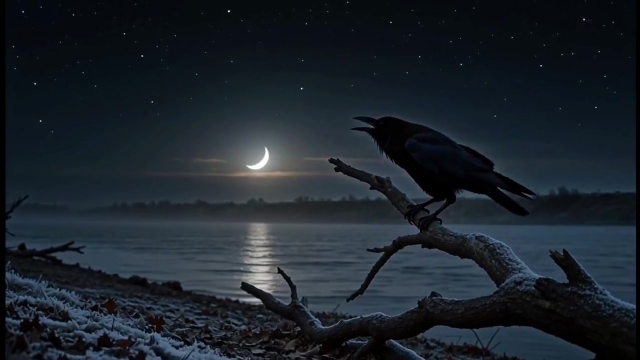} &
\includegraphics[width=.315\textwidth]{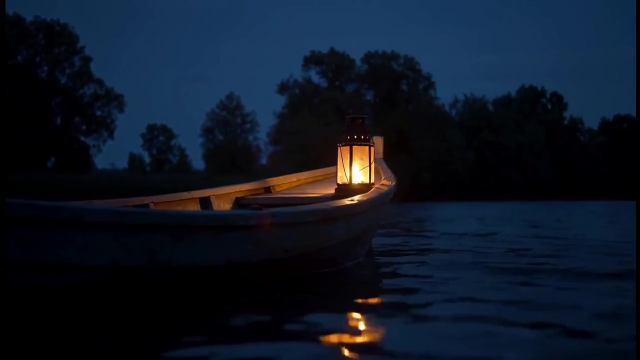} &
\includegraphics[width=.315\textwidth]{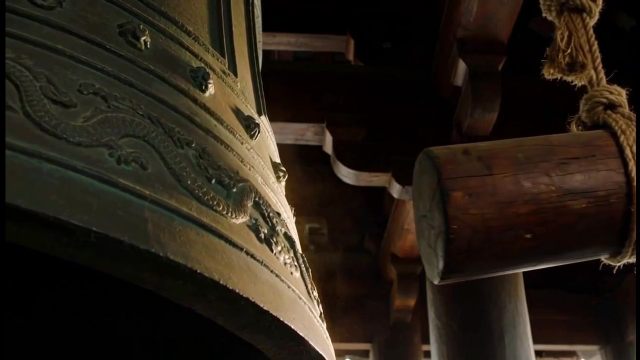} \\
\scriptsize (g) Moonset and crow calls &
\scriptsize (h) Fishing lights &
\scriptsize (i) The temple bell
\end{tabular}
\caption{Selected frames from the generated teaching video for Zhang Ji's
\emph{Mooring by Maple Bridge}. Frames (a)--(f) move from the contextual
narrative to the riverside setting; frames (g)--(i) visualize the moonset
and crow calls, fishing lights, and the bell at Hanshan Temple.}
\label{fig:case_fengqiao}
\end{figure*}

\begin{figure*}[!t]
\centering
\setlength{\tabcolsep}{2pt}
\renewcommand{\arraystretch}{0.85}
\begin{tabular}{ccc}
\includegraphics[width=.315\textwidth]{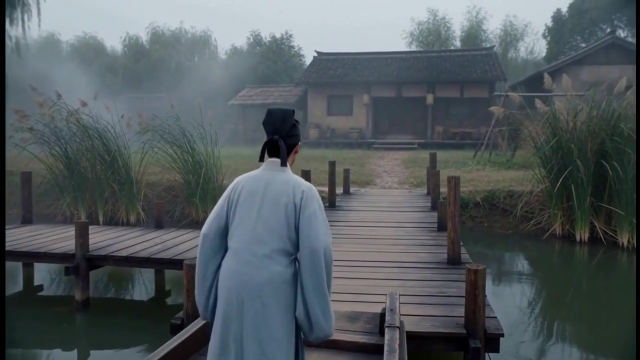} &
\includegraphics[width=.315\textwidth]{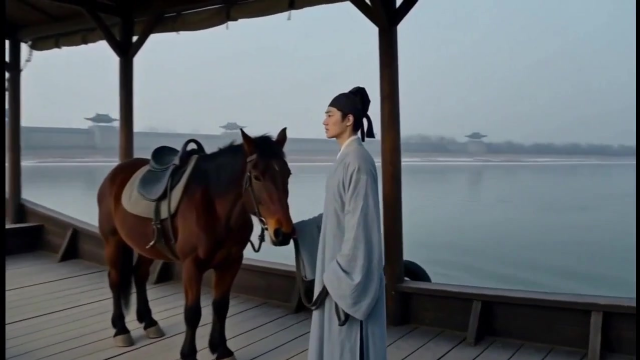} &
\includegraphics[width=.315\textwidth]{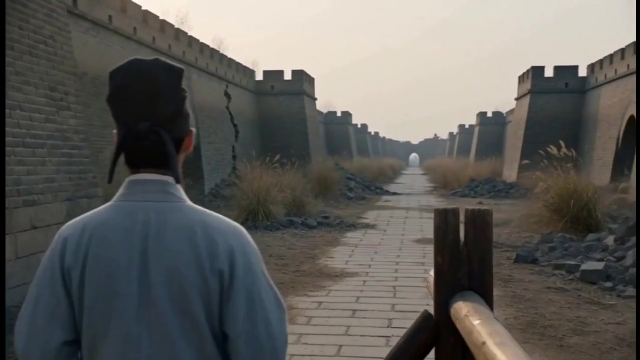} \\
\scriptsize (a) Early displacement &
\scriptsize (b) Journey to Yangzhou &
\scriptsize (c) Arrival at the city \\[3pt]
\includegraphics[width=.315\textwidth]{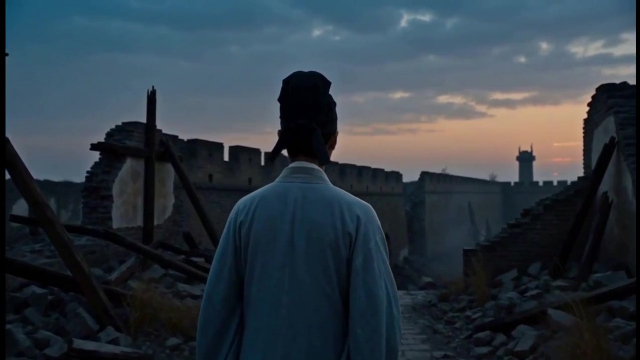} &
\includegraphics[width=.315\textwidth]{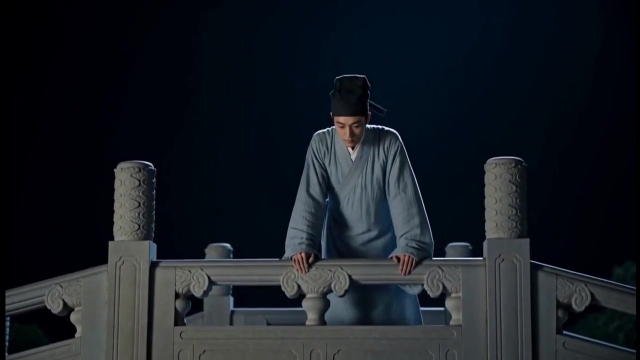} &
\includegraphics[width=.315\textwidth]{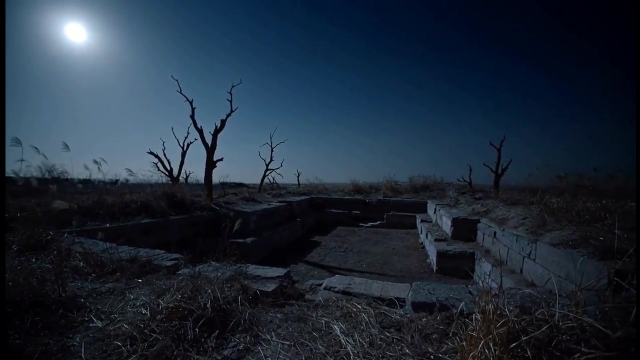} \\
\scriptsize (d) Postwar ruins &
\scriptsize (e) Bridge-side reflection &
\scriptsize (f) Moonlit desolation \\[3pt]
\includegraphics[width=.315\textwidth]{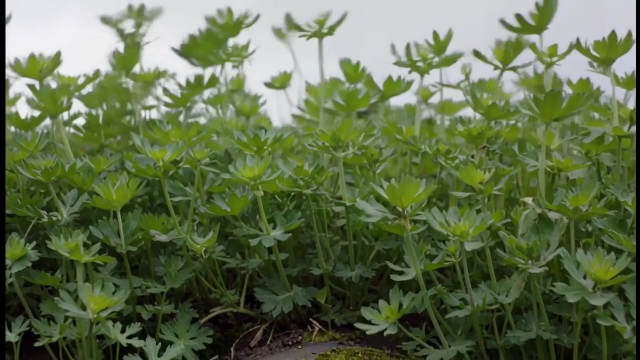} &
\includegraphics[width=.315\textwidth]{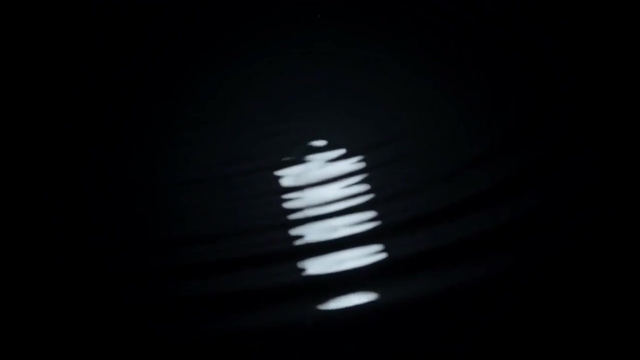} &
\includegraphics[width=.315\textwidth]{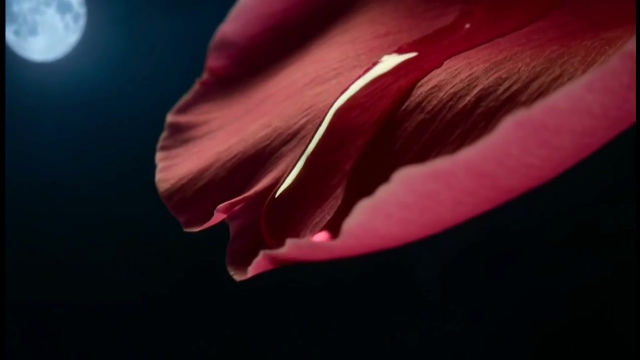} \\
\scriptsize (g) Wild mustard &
\scriptsize (h) The cold moon &
\scriptsize (i) Bridge-side peonies
\end{tabular}
\caption{Selected frames from the generated teaching video for Jiang Kui's
\emph{Yangzhou Man}. Frames (a)--(f) establish the journey to Yangzhou and
the contrast between the remembered city and its postwar ruins; frames
(g)--(i) visualize wild mustard, the cold moon, and bridge-side peonies.}
\label{fig:case_yangzhou}
\end{figure*}

\noindent\textbf{Gemini-based evaluation.}
The narration receives 0.88 for Required Information Coverage; 1.00 for
Metaphor Source Identification, Target Identification, Mapping Justification,
and Contextual Grounding; and 0.80 for Interpretive Calibration. Narrative
Coherence and Storytelling--Teaching Transition are both rated 5/5.

\paragraph{Analysis.}
The generated background story frames the poem through examination success,
displacement, and travel before converging on the nocturnal boat scene. This
historical framing is presented as a narrative interpretation rather than as
a definitive account of the poem's occasion. The line-by-line reading follows
the poem's sensory progression from moonset and crow calls to riverside
fishing lights and the distant temple bell. The visual sequence makes the
movement from nearby images to distant sound explicit without requiring
additional on-screen exposition.

%\clearpage

\subsection{\emph{Yangzhou Man}: Challenging Song-Ci Example}
\label{app:case_yangzhou}

\noindent\textbf{Gemini-based evaluation.}
The narration receives 1.00 for Required Information Coverage and all five
metaphor criteria (Source Identification, Target Identification, Mapping
Justification, Contextual Grounding, and Interpretive Calibration). Narrative
Coherence is rated 5/5, and Storytelling--Teaching Transition is rated 4/5.

\paragraph{Analysis.}
This example combines historical context with the spatial and intertextual
structure of a longer Song-ci text. The background story contrasts the city's
remembered prosperity with its postwar desolation. The line-by-line reading
then uses wild mustard, the cold moon, and bridge-side peonies as recurring
visual anchors for the contrast between past and present. The generated
narration interprets this contrast as joining personal displacement with
lament for historical decline; this is an interpretive framing rather than a
claim of a single definitive meaning.

% \section{Prompt}
% \label{sec:prompt}
\clearpage
\onecolumn

\section{Prompt Templates}
\label{sec:prompt}

We present prompt templates organized by the
planning and realization stages introduced in Section~\ref{sec:method}. Gold annotations used for evaluation are not exposed to any generation stage.

% ------------------------------------------------------------
% Semantic Planning
% ------------------------------------------------------------

\begin{promptcard}
{Semantic Planning ($\Pi$)}

\promptlabel{System message}

\emph{You are the Knowledge Planner for classical Chinese poetry.
Recover the contextual and interpretive information needed to explain the
poem and organize it into a semantic Director Graph for downstream
generation. Distinguish supported evidence from uncertain model knowledge,
and do not introduce unsupported facts. Return structured output only.}

\tcblower

\promptlabel{Instructions}

\begin{enumerate}[leftmargin=1.35em,itemsep=1pt,topsep=2pt]

\item Identify the historical, biographical, situational, and interpretive
information needed to understand the poem beyond its literal wording.

\item Organize the available information into contextual events or states,
their relevant relations, figurative mappings, and global visual constraints.

\item Treat the poem and metadata as direct input evidence. When external
background material is provided, preserve its provenance and use it without
altering its claims.

\item When external material is insufficient or absent, model-recalled
background knowledge may supplement the context. Such recalled claims must
remain explicitly unverified and cannot be upgraded to documented fact or
scholarly consensus.

\item Associate factual and interpretive claims with their supporting evidence
and assign an epistemic status no stronger than that evidence permits.

\item Do not infer a historical writing scene, authorial intention, or
biographical event solely from imagery or actions described in the poem.

\item Preserve uncertainty for disputed or weakly supported interpretations.
Do not invent sources, quotations, historical records, or claims of scholarly
consensus.

\item Cover the poem as a whole, but do not write the final background story,
narration, character assets, shot descriptions, or video prompts.

\end{enumerate}

\promptlabel{Inserted payloads}

\promptfield{POEM_RECORD}:
\promptfield{<POEM_AND_METADATA>};

\promptfield{BACKGROUND_MATERIAL}:
\promptfield{<OPTIONAL_PROVIDED_BACKGROUND_OR_EMPTY>}.

\end{promptcard}

% ------------------------------------------------------------
% Background Story
% ------------------------------------------------------------

\begin{promptcard}
{Background Story ($\phi_1$)}

\promptlabel{System message}

\emph{You are a contextual storyteller for classical Chinese poetry.
Using only the relevant context and evidence in the Director Graph, write a
coherent background story that makes the poem's implicit historical or
situational context understandable. Do not invent unsupported events,
dialogue, or authorial psychology.}

\tcblower

\promptlabel{Instructions}

\begin{enumerate}[leftmargin=1.35em,itemsep=1pt,topsep=2pt]

\item Select the contextual thread most appropriate to the work, such as a
historical event, biographical episode, spatial situation, object development,
or another supported progression.

\item Present the background as one continuous story rather than a list of
facts. Preserve the temporal and causal relations supported by the Director
Graph without imposing a fixed dramatic structure.

\item Use only concrete details licensed by the supplied context and evidence.
Do not invent dialogue, motives, inner monologue, or unsupported sensory
detail.

\item If the historical writing situation is unknown, do not reconstruct the
author's physical presence from the poem itself.

\item Preserve uncertainty naturally when the evidence is incomplete or an
interpretation is disputed.

\item End with a bridge that connects the contextual story to the poem without
asserting a single definitive interpretation.

\end{enumerate}

\promptlabel{Inserted payload}

\promptfield{STORY_CONTEXT}:
\promptfield{<RELEVANT_CONTEXT_AND_EVIDENCE_FROM_DIRECTOR_GRAPH>}.

\end{promptcard}

% ------------------------------------------------------------
% Narration Segmentation
% ------------------------------------------------------------

\begin{promptcard}
{Narration Segmentation ($\phi_2$)}

\promptlabel{System message}

\emph{You are a narration writer. Convert the locked background story into
connected spoken segments while preserving its factual and interpretive
content. For each segment, identify the visual events required to communicate
the narration and assign a provisional duration for downstream audiovisual
planning.}

\tcblower

\promptlabel{Instructions}

\begin{enumerate}[leftmargin=1.35em,itemsep=1pt,topsep=2pt]

\item Rewrite the background story as a sequence of natural spoken narration
segments whose order follows the underlying contextual progression.

\item Preserve the factual and interpretive boundaries established by the
background-story stage. Do not add new names, dates, places, events, motives,
or historical claims.

\item For each narration segment, identify the concrete visual events needed
to communicate its content. These events describe what must happen visually,
not how the camera should film it.

\item Assign each narration segment a provisional duration that can later be
revised after speech synthesis.

\item Include only people who are visually present when character identity
matters for downstream generation.

\item Preserve uncertainty in disputed interpretations rather than converting
them into definite claims.

\item Do not specify camera motion, shot composition, video-model instructions,
or final editing decisions.

\end{enumerate}

\promptlabel{Inserted payload}

\promptfield{LOCKED_BACKGROUND_STORY}:
\promptfield{<BACKGROUND_STORY_AND_RELEVANT_GRAPH_VIEW>}.

\end{promptcard}

% ------------------------------------------------------------
% Line-by-Line Reading
% ------------------------------------------------------------

\begin{promptcard}
{Line-by-Line Reading ($\phi_3$)}

\promptlabel{System message}

\emph{You are an explanatory narration writer for classical Chinese poetry.
Explain the poem in source order using the poem-centered interpretation and
the background context needed to connect that interpretation to the preceding
story. Ground each explanation in the text and preserve uncertainty around
contested readings.}

\tcblower

\promptlabel{Instructions}

\begin{enumerate}[leftmargin=1.35em,itemsep=1pt,topsep=2pt]

\item Follow the poem in source order and cover the textual units needed for a
complete explanation.

\item Explain important words, objects, actions, syntax, imagery, allusions,
structural turns, and figurative relations where they contribute to the
meaning of the poem.

\item For figurative expressions, make clear the relevant source domain,
target interpretation, and the relation that connects them when supported by
the available evidence.

\item Connect the close reading to relevant background context when that
context helps explain the poem, but do not simply repeat the preceding
background narration.

\item Ground interpretive claims in specific textual or contextual evidence.
Avoid generic statements about mood, beauty, or emotional effect when a more
concrete explanation is available.

\item Distinguish supported interpretations from plausible or disputed ones,
using appropriately qualified language where necessary.

\item Do not introduce new historical facts or modify information fixed by the
Director Graph and background-story stage.

\end{enumerate}

\promptlabel{Inserted payloads}

\promptfield{POEM}:
\promptfield{<POEM_TEXT_AND_LINE_IDS>};

\promptfield{READING_CONTEXT}:
\promptfield{<INTERPRETATION_AND_RELEVANT_BACKGROUND_CONTEXT>}.

\end{promptcard}

% ------------------------------------------------------------
% Character Assets
% ------------------------------------------------------------

\begin{promptcard}
{Character Assets ($\phi_4$)}

\promptlabel{System message}

\emph{You are a character-asset designer for historical video generation.
Create reusable visual specifications for people who recur across multiple
visual units. The specifications are used to maintain identity consistency,
not to claim historically verified portrait reconstruction.}

\tcblower

\promptlabel{Instructions}

\begin{enumerate}[leftmargin=1.35em,itemsep=1pt,topsep=2pt]

\item Create a reusable character asset only for people who recur across
multiple visual units and therefore require cross-shot identity consistency.

\item For each recurring person, provide a synthetic reference-image
specification together with stable appearance anchors that can be reused by
the video generator.

\item Preserve stable identity cues across appearances while allowing
contextually motivated changes such as age, clothing, or physical condition.

\item Use only appearance information supported by the supplied context.
When historical appearance is unknown, use a synthetic design rather than
presenting the generated appearance as a reconstruction of the real person.

\item Avoid recognizable modern or celebrity identities.

\item When an interpretation requires a figure to remain anonymous or
ambiguous, preserve that ambiguity rather than assigning a specific identity.

\item Do not alter the narration, event structure, or interpretation
established by earlier stages.

\end{enumerate}

\promptlabel{Inserted payload}

\promptfield{CHARACTER_CONTEXT}:
\promptfield{<GRAPH_VIEW_AND_RECURRING_VISUAL_ENTITIES>}.

\end{promptcard}

% ------------------------------------------------------------
% Visual Direction
% ------------------------------------------------------------

\begin{promptcard}
{Visual Direction ($\phi_5$)}

\promptlabel{System message}

\emph{You are the visual director for a classical-poetry video. Translate the
locked narration and event structure into shot-level visual descriptions.
Preserve the meaning fixed by earlier stages, maintain recurring-character
consistency, and organize the resulting shots for narration-aligned
audiovisual assembly.}

\tcblower

\promptlabel{Instructions}

\begin{enumerate}[leftmargin=1.35em,itemsep=1pt,topsep=2pt]

\item Preserve the order and semantic content of the locked narration and
event descriptions. Do not rewrite narration or historical claims.

\item Write each shot as a self-contained visual-generation description that
specifies the scene, people, setting, action, and intended ending state
without changing the event it represents.

\item Use the supplied character assets whenever recurring people appear so
that identity remains consistent across shots.

\item Respect the global visual constraints in the Director Graph, including
historical setting and any forbidden realizations.

\item For figurative or disputed interpretations, preserve the intended
semantic relation without forcing an unsupported literal realization.

\item Group the resulting shots into edit groups aligned with their
corresponding narration segments. Each group should preserve the order of its
visual events and provide provisional shot allocations for later timing
adjustment.

\item Treat these durations as provisional. Final shot timing is determined
after speech synthesis using the measured narration duration.

\item Do not add subtitles, readable modern text, logos, watermarks,
anachronistic objects, or other content that contradicts the supplied visual
constraints.

\end{enumerate}

\promptlabel{Inserted payloads}

\promptfield{GLOBAL_VISUAL_CONTEXT}:
\promptfield{<DIRECTOR_GRAPH_VIEW_AND_CHARACTER_ASSETS>};

\promptfield{LOCKED_NARRATION_AND_EVENTS}:
\promptfield{<ORDERED_NARRATION_AND_EVENT_STRUCTURE>}.

\end{promptcard}

\end{document}